\documentclass[10pt,twocolumn,letterpaper]{article}

\usepackage[pagenumbers]{cvpr}
\usepackage{multirow}

\definecolor{cvprblue}{rgb}{0.21,0.49,0.74}
\usepackage[pagebackref,breaklinks,colorlinks,allcolors=cvprblue]{hyperref}

\def\confName{CVPR}
\def\confYear{2027}

\newcommand{\method}{MotifFlow}
\newcommand{\adflux}{AD-FLUX}
\newcommand{\inj}{\mathrm{inj}}
\newcommand{\self}{\mathrm{self}}

\title{MotifFlow: Training-Free, Self-Contained Style Transfer\\ from Color to Motif in Rectified Flow Transformers}

\author{Chun Xie \qquad Itaru Kitahara\\
University of Tsukuba\\
{\tt\small \{xie.chun,\,kitahara.itaru\}@image.iit.tsukuba.ac.jp}
}

\begin{document}
\maketitle
\begin{abstract}
Expressive style transfer must reproduce a painting's palette, brushwork, and characteristic motifs while retaining the content image's structure.
Attention injection is effective in U-Net-based diffusion models, but adapting it to FLUX often transfers color without capturing the style's distinctive brushwork.
Our analysis across backbones links this gap to how injected features are integrated across neighboring spatial locations.
We introduce \method{}, a training-free, self-contained framework that combines style injection with style-alignment guidance in FLUX.
Injection supplies style features during generation but does not explicitly encourage the model's native attention output to align with them. We therefore use the discrepancy between native and injected attention outputs to guide latent-space updates, encouraging the evolving image to develop the style's distinctive brushwork and motifs.
Style encoding, guidance, and synthesis all use the same frozen backbone, without auxiliary image encoders or trained adapters.
A single strength parameter controls the transition from color transfer to pronounced brushwork and motifs, with additional support for regional stylization and style mixing.
Experiments on 1,000 content--style pairs show that \method{} better matches the local textures of the style image than baseline methods and produces more consistent stylization than injection alone. The same design extends to FLUX.2.
\end{abstract}

\section{Introduction}
\label{sec:intro}

\begin{figure*}[t]
\centering
\includegraphics[width=0.9\linewidth]{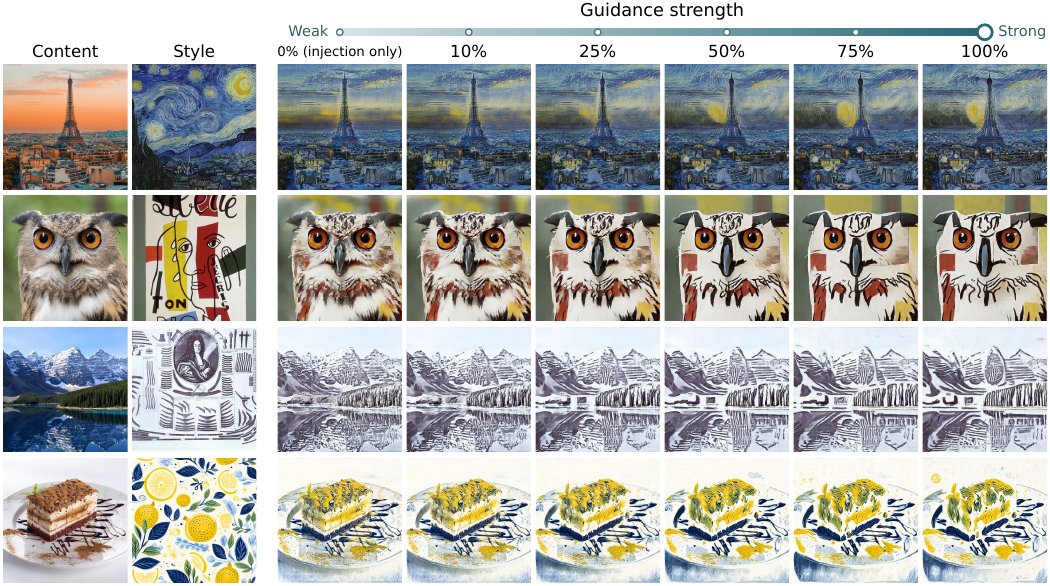}
\caption{\textbf{From color to motif.} Each row shows the same content--style pair at six guidance strengths from 0\% to 100\%. Zero guidance uses attention injection alone. Increasing guidance develops style-specific appearance: swirling painted strokes around the Eiffel Tower, graphic outlines on the owl, engraved line patterns across the mountain lake, and citrus and leaf motifs in the cake scene. The content and style images are shown in the first two columns.}
\label{fig:teaser}
\end{figure*}

Style transfer must balance changes in appearance with preservation of content. A longstanding goal is to separate stylistic appearance from the content depicted in the style image~\cite{gatys2016image,wang2024instantstyle}. Yet motifs such as the swirling forms in the sky of Van Gogh's \emph{The Starry Night} and recurring elements in decorative patterns resist this distinction. Although these recognizable forms may be regarded as content, they are also essential elements of the style being transferred. Expressive transfer therefore calls for control over a spectrum from palette and brushwork to these defining motifs.

Attention injection provides a direct way to introduce the style image's appearance. StyleID~\cite{chung2024styleid} replaces attention keys and values with style features in a frozen U-Net-based diffusion model, using query blending to preserve content structure and attention sharpening to strengthen stylization. Together with related feature-sharing methods~\cite{hertz2024stylealigned,alaluf2024crossimage}, this establishes attention as a useful interface for training-free style transfer.

We investigate whether attention injection can achieve similarly expressive stylization in FLUX.1~\cite{bfl2024flux}, a rectified-flow model~\cite{liu2023rectifiedflow} with a transformer backbone~\cite{peebles2023dit,esser2024sd3}. FLUX-based systems such as USO~\cite{wu2025uso} and Redux~\cite{bfl2024redux} encode conditioning images through auxiliary image encoders and trained projection layers. Motivated by StyleID's self-contained design, we seek to perform style transfer using only the pretrained FLUX backbone.

Direct attention injection behaves differently on this backbone. FLUX's rotary position embedding (RoPE) couples feature matching to spatial coordinates, so cross-image injection can pull the output toward the style image's composition. Removing RoPE at the injection blocks limits this copying and transfers the palette, but distinctive brushwork remains weak. Sharpening mostly boosts contrast and can introduce grid artifacts. Our cross-backbone analysis links these limitations to neighboring feature integration: U-Nets retain convolutional mixing, whereas removing RoPE weakens the local attention bias at FLUX's injection blocks.

To address the weak transfer of distinctive brushwork while retaining position-free injection, we introduce \method{}, a training-free, self-contained framework with style alignment guidance. The injected attention output supplies style features during generation and also serves as an alignment target. Alongside it, we compute a native attention output from the evolving image's own features. Injection alone does not explicitly encourage this native output to match the injected one. We therefore use their discrepancy to guide latent-space updates, encouraging the evolving image to develop the style's distinctive brushwork and motifs. A content-query anchor preserves structure, while a lightweight grain loss complements alignment by matching local texture statistics. By scaling these latent guidance updates with a single strength parameter, we control stylization from palette transfer to pronounced brushwork and motifs. Beyond global strength, the framework also supports regional stylization, style-detail selection, and multi-style composition.

Style encoding, guidance, and synthesis all use the same frozen backbone. This self-contained design enables expressive style transfer using the generator's existing visual representations, without an auxiliary image encoder or learned projection module.

Our contributions are:
\begin{itemize}
\item \textbf{Cross-architecture analysis of attention injection.} We link the limitations of direct injection on FLUX to neighboring feature integration, motivating the use of latent guidance to produce coherent brushwork.
\item \textbf{Self-contained style transfer.} \method{} couples position-free style injection with style alignment guidance, enabling expressive stylization using only the frozen backbone's own representations.
\item \textbf{Control from color to motif.} A single strength parameter provides continuous control from palette transfer to distinctive brushwork and motifs.
\end{itemize}

Experiments show the lowest texture distance among the compared methods and a reduction in style-transfer failures (CSD below $0.30$) from 29.6\% for injection alone to 1.2\%. At matched perceptual change from the content image (LPIPS), \method{} also achieves stronger style resemblance and greater local texture similarity than Attention Distillation~\cite{zhou2025attentiondistillation} adapted to FLUX, while running approximately $8.2\times$ faster. Results on FLUX.2 further demonstrate that the design extends beyond the primary backbone.

\section{Related Work}
\label{sec:related}

\paragraph{Image Conditioning for Style Transfer.}
Early neural style transfer matches feature statistics to reproduce the style image's appearance~\cite{gatys2016image,huang2017adain,li2017universal}. With pretrained generative models, image conditioning offers another route: IP-Adapter~\cite{ye2023ipadapter} learns to introduce image-encoder features into diffusion sampling, while InstantStyle~\cite{wang2024instantstyle} separates style from content in these features and selects where to inject them. InstantStyle-Plus~\cite{wang2024instantstyleplus} further incorporates content-preserving controls. On FLUX, Redux~\cite{bfl2024redux} and USO~\cite{wu2025uso} also use auxiliary image encoders and learned conditioning modules. OmniStyle~\cite{wang2025omnistyle} trains a transformer for style transfer using paired data. In contrast, our self-contained approach uses the frozen backbone's own features for both style representation and guidance.

\paragraph{Attention Sharing and Injection.}
The generative backbone's own attention features also provide a style representation. StyleAligned~\cite{hertz2024stylealigned} shares attention across generated images to maintain a common style. Cross-Image Attention~\cite{alaluf2024crossimage} transfers appearance by matching features across images, and Eye-for-an-eye~\cite{go2026eyeforaneye} rearranges style features using dense semantic correspondence. For content-image stylization, StyleID~\cite{chung2024styleid} combines style key--value injection with content-query preservation and attention sharpening. These methods introduce the style image's appearance through the forward attention computation. Their effectiveness therefore depends on how the backbone processes the supplied features.

\paragraph{Optimization for Appearance Transfer.}
Style features can also define an objective for updating the generated image. Gradient-based guidance provides a general way to steer diffusion sampling toward such objectives~\cite{yu2023freedom,bansal2023universal}. Attention Distillation~\cite{zhou2025attentiondistillation} matches attention outputs to transfer visual characteristics through latent optimization or guided sampling. StyleGallery~\cite{he2026stylegallery} performs regional appearance matching using semantic correspondences that include DINOv2~\cite{oquab2024dinov2} features. Inspired by attention-output matching, we use the active injection output as the alignment target during sampling, coupling forward style injection with latent guidance.

\paragraph{Style Transfer on Flow Transformers.}
Recent work adapts style transfer to rectified-flow transformers through trajectory design and attention control. Dual-RF~\cite{deng2025dualrf} combines content and style velocity fields with attention injection to avoid inversion. Untwisting RoPE~\cite{mikaeili2026untwisting} identifies positional bias as a source of style image copying in shared attention and modulates style keys by frequency to support stylized generation. SuRe~\cite{tang2025sure} suppresses semantic leakage by shuffling style values and progressively restoring key--value correspondence during latent optimization, with results on both U-Net and flow backbones. To understand why direct injection leaves style-specific detail weak on FLUX, our analysis examines how attention sharpening and positional encoding interact with local feature integration.

\section{Motivation}
\label{sec:motivation}

\begin{figure}[t]
\centering
\includegraphics[width=0.88\linewidth]{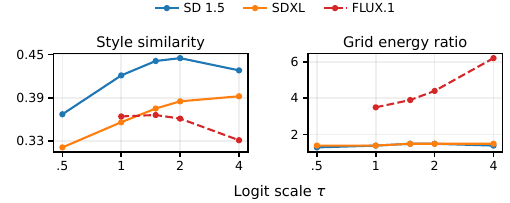}
\par\smallskip
\includegraphics[width=0.88\linewidth]{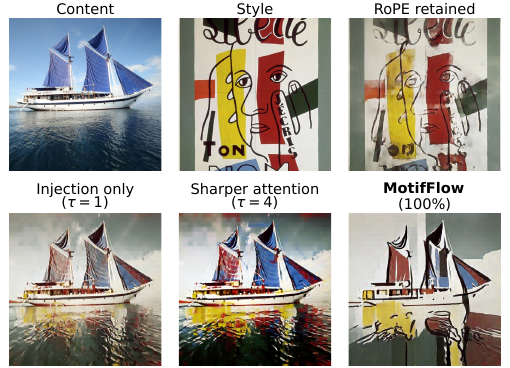}
\caption{\textbf{Limits of direct injection.} Sharpening strengthens U-Net stylization but increases grid artifacts on FLUX. Sharper FLUX attention mainly boosts contrast, while retaining RoPE biases the output toward the style image's layout. MotifFlow transfers the style while preserving content structure.}
\label{fig:motivation}
\end{figure}

\paragraph{Adapting Attention Injection to FLUX.}
StyleID~\cite{chung2024styleid} transfers style within a pretrained U-Net diffusion model by replacing attention keys and values with style features. Applying this idea to FLUX introduces a positional constraint: rotary position embeddings (RoPE)~\cite{su2024roformer} make attention depend on the spatial coordinates of queries and keys. When queries from the evolving image attend to style keys, RoPE makes their matching depend on the tokens' relative positions. This pulls the output toward the style image's composition (\cref{fig:motivation} and Appendix~A.1). We therefore omit RoPE at the injection blocks to match style features without enforcing spatial correspondence between the two images. This limits copying of the style image's layout while retaining the style image's influence on appearance.

\paragraph{Why Sharpening Is Insufficient.}
Position-free injection preserves the content structure and transfers broad appearance, including the style palette and some texture, but remains weak at expressing the style image's distinctive brushwork (\cref{fig:motivation} and Appendix~A.2). A natural next step is StyleID's attention sharpening: increasing the logit multiplier $\tau$ concentrates attention weights on fewer positions in the style image. On U-Net diffusion models, increasing attention sharpness strengthens stylization. On FLUX, increasing $\tau$ mostly boosts color contrast and can introduce grid artifacts, rather than developing the style image's characteristic brushwork (\cref{fig:motivation} and Appendix~A.3).

Sharpening concentrates attention on selected style features, but coherent brushwork also depends on how the backbone integrates these features across neighboring positions. U-Net convolutions mix neighboring features independently of where attention retrieves them from. In FLUX, spatial interactions are mediated by attention, with RoPE encoding relative positional information. Our attention measurements show that removing RoPE reduces attention to neighboring image positions at the injection blocks (Appendix~A.4). As a result, sharpening can amplify differences between neighboring token outputs, making the token grid visible in the generated image. This helps explain why sharper style matching in FLUX does not produce the same improvement in coherent brushwork.

\paragraph{From Injection to Style Alignment.}
We therefore seek to develop style-specific appearance while retaining position-free injection. The injected attention output supplies style features during generation, but injection alone does not explicitly encourage the native attention output to match it. To provide this constraint, we use the injected output as an alignment target and update the latent to reduce the discrepancy between the two outputs, while keeping injection active. This style alignment guidance uses the same frozen backbone for style injection and latent optimization, preserving a self-contained transfer process.

\section{Method}
\label{sec:method}

\begin{figure*}[t]
\centering
\includegraphics[width=0.9\linewidth]{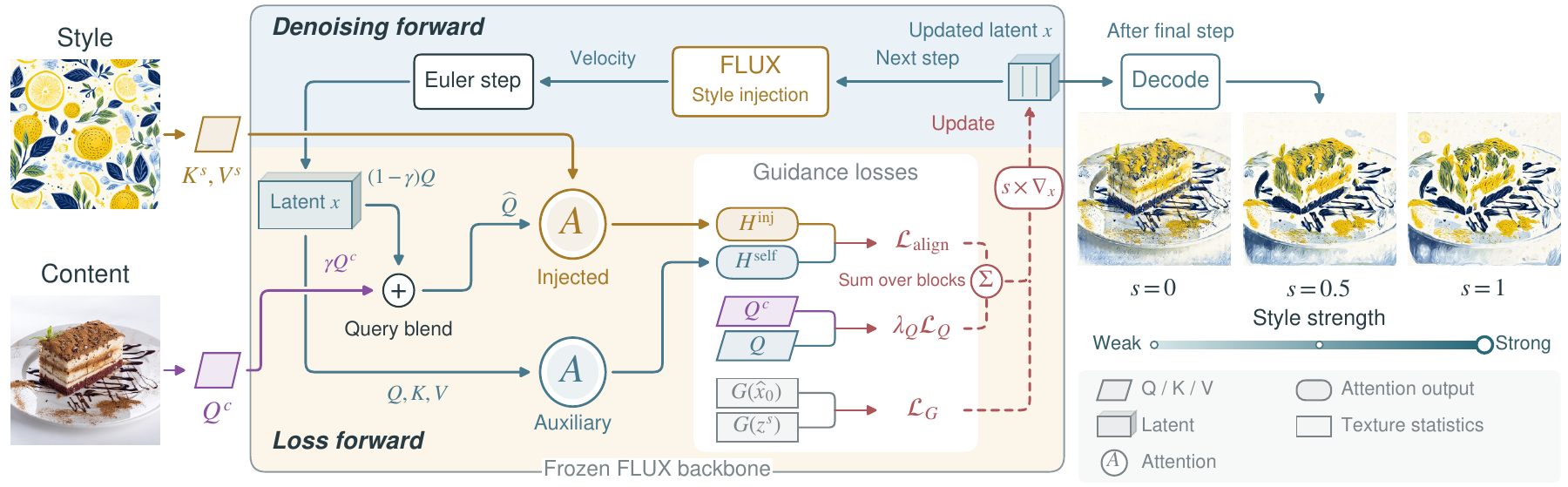}
\caption{\textbf{Style injection and latent alignment.} Each sampling step uses a denoising forward followed by a differentiable loss forward. Both inject style features into selected blocks without RoPE. In the loss forward, cached content queries blend with live queries using $\gamma$, and the injected attention output $H^{\mathrm{inj}}$ provides the alignment target for the auxiliary output $H^{\mathrm{self}}$. Alignment and query losses accumulate across selected blocks, while grain guidance compares latent texture statistics. Their gradients update $x$ for the next denoising step, with strength controlled by $s$. Results show independent runs at $s=0$, $0.5$, and $1$. Text-token paths are omitted.}
\label{fig:method}
\end{figure*}

Given content image $I^c$ and style image $I^s$, \method{} generates a stylized image with a frozen rectified-flow model. At each sampling step, position-free injection supplies style features to the forward pass, and style alignment guidance uses the resulting attention output to update the latent (\cref{fig:method}). We complement style alignment with grain guidance, which matches local texture statistics. A single strength parameter $s\in[0,1]$ scales both latent updates.

\subsection{Style Feature Injection}
\label{sec:method:inject}

Let $x$ denote the sampling latent and $\sigma$ the noise level, which decreases from 1 to 0 during generation. We invert both inputs and cache the content queries $Q^c$ and style keys and values $(K^s,V^s)$ for each sampling timestep and selected transformer block $\ell\in\mathcal B$. Generation starts from the inverted content noise, with its channel statistics matched to the style noise following StyleID~\cite{chung2024styleid}. As motivated in \cref{sec:motivation}, we omit RoPE in $\mathcal B$ during inversion, sampling, and guidance. Other blocks retain RoPE.

At an injection block, let $(Q,K,V)$ denote the live image queries, keys, and values, and $(K^t,V^t)$ the current text keys and values. Define $A(Q,K,V)=\operatorname{softmax}(QK^\top/\sqrt d)V$, where $d$ is the head dimension. To retain content structure, we blend the live queries with the cached content queries before reading the style features:
\begin{align}
\widehat Q &= \gamma Q^c+(1-\gamma)Q, \label{eq:qblend}\\
H^{\inj} &= A(\widehat Q,[K^t;K^s],[V^t;V^s]). \label{eq:inject}
\end{align}
Brackets denote concatenation along the token dimension. The injected attention output $H^{\inj}$ replaces the image-token attention output, while text-token outputs retain ordinary attention. We omit block and timestep indices here for clarity.

\subsection{Style Alignment Guidance}
\label{sec:method:align}

To make the current image express the injected appearance, we compare $H^{\inj}$ with an auxiliary attention output computed from its own features:
\begin{equation}
H^{\self}_{\ell}=A(Q_{\ell},[K^t_{\ell};K_{\ell}],[V^t_{\ell};V_{\ell}]).
\label{eq:self}
\end{equation}
Only $H^{\inj}$ continues the forward pass. The auxiliary $H^{\self}$ is used to measure the alignment loss,
\begin{equation}
\mathcal L_{\mathrm{align}}=\sum_{\ell\in\mathcal B}\operatorname{MSE}(H^{\self}_{\ell},H^{\inj}_{\ell}).
\label{eq:lalign}
\end{equation}
The mean squared error (MSE) averages over image tokens, heads, and channels. This attention-output objective is inspired by Attention Distillation~\cite{zhou2025attentiondistillation}, with the target here supplied by the active injection path.

To preserve content structure during optimization, we anchor the live queries to their cached content counterparts with weight $\lambda_Q$:
\begin{equation}
\mathcal L=\mathcal L_{\mathrm{align}}+
\lambda_Q\sum_{\ell\in\mathcal B}\operatorname{MAE}(Q_\ell,Q^c_\ell).
\label{eq:total}
\end{equation}
The mean absolute error (MAE) averages over the same dimensions as MSE. We backpropagate the loss through the frozen transformer to update the latent.

After each Euler step, we perform an additional differentiable forward pass at the updated latent and its corresponding noise level to compute $\mathcal L$. We then use the loss gradient to update the latent:
\begin{equation}
x\leftarrow x-s\,\eta_{\mathrm{align}}\,\bar g,
\qquad g=\nabla_x\mathcal L.
\label{eq:step}
\end{equation}
Here, $\eta_{\mathrm{align}}$ is the step size and $\bar g$ is the clipped, normalized gradient. Token-wise clipping suppresses isolated saturated speckles caused by unusually large local updates, while normalization stabilizes the update scale (Appendix~B.2). Alignment guidance operates throughout sampling, with injection active in every forward evaluation.

\subsection{Grain Guidance and Style Strength}
\label{sec:method:extra}

Alignment guidance constrains transformer features. Grain guidance complements it by matching Gram matrices of high-pass neighborhood features from the current clean latent estimate and the style latent (loss and feature construction: Appendix~B.3). This texture loss requires no additional network pass. Its gradient contributes to latent guidance at intermediate noise levels.

The strength parameter $s$ scales both guidance step sizes (default settings: Appendix~B). At $s=0$, generation uses injection alone. Increasing $s$ strengthens both latent updates while leaving injection unchanged, providing the progression from color to brushwork and motifs shown in \cref{fig:teaser}.

\subsection{Region Selection and Style Mixing}
\label{sec:method:control}

Beyond global strength, we control the spatial extent of additional stylization by applying a mask derived from the content image to both latent guidance updates. We obtain the mask from the difference between the backbone's text-to-image attention maps for two contrasting descriptions, such as the subject and background. No external segmentation model is needed.

Applying the same procedure to the style image allows us to select specific elements, such as lemon or leaf motifs. The resulting mask restricts the style keys and values used to compute the alignment target, directing alignment guidance toward the selected elements. Neither content-region masking nor style-element selection modifies the forward injection path.

To combine two styles, we concatenate their cached keys and values and control their relative preference through attention biases. A content mask favors the first style where the mask is high and the second where it is low, assigning styles by region. Global mixing instead uses constant weights $1-w$ and $w$ at every image location, with $w\in[0,1]$. Both modes apply these biases to forward injection and the alignment target. Appendix~C illustrates where these controls act and gives the mask extraction and attention-weighting details.

\section{Experiments}
\label{sec:exp}

\subsection{Experimental Setup}
\label{sec:exp:setup}

Our benchmark combines 40 content images with 25 style images to form 1,000 content--style pairs. Content spans people, animals, architecture, landscapes, and objects, while styles range from oil painting and ink to mosaic and graphic design. Unless otherwise stated, we evaluate \method{} using frozen FLUX.1-dev with 20 Euler steps at $512\times512$, an empty prompt, and the same parameter settings across styles (implementation: Appendix~B).

We compare with USO~\cite{wu2025uso} and Redux~\cite{bfl2024redux}, which provide image conditioning on FLUX, and with the style-transfer methods StyleID~\cite{chung2024styleid}, StyleGallery~\cite{he2026stylegallery}, and Attention Distillation (AD)~\cite{zhou2025attentiondistillation} on SD\,1.5 (configurations: Appendix~D.2). Redux receives the style image through its adapter and the content image through image-to-image initialization. Injection alone provides the starting point for assessing the benefit of our guidance.

\paragraph{Evaluation Metrics.}
\textbf{LPIPS}~\cite{zhang2018lpips} measures perceptual change from the content image. \textbf{Structure} measures structural deviation from the content image using the mean absolute difference between the output and content images' DINO self-similarity matrices~\cite{caron2021dino}. \textbf{CSD}~\cite{somepalli2024csd} measures overall style resemblance. We quantify insufficient style transfer using the \textbf{Failure} rate, defined as the percentage of outputs with CSD below $0.30$. \textbf{Color} compares the CIELAB pixel-color distributions of the output and style image using sliced Wasserstein-2 distance. \textbf{Texture} compares VGG-19~\cite{simonyan2015very} Gram matrices of their high-pass-filtered grayscale images using mean squared error. These networks are used only for evaluation. See Appendix~D.3 for metric definitions and additional style statistics.

\subsection{Style Transfer Quality}
\label{sec:exp:main}

\begin{figure*}[t]
\centering
\includegraphics[width=0.9\linewidth]{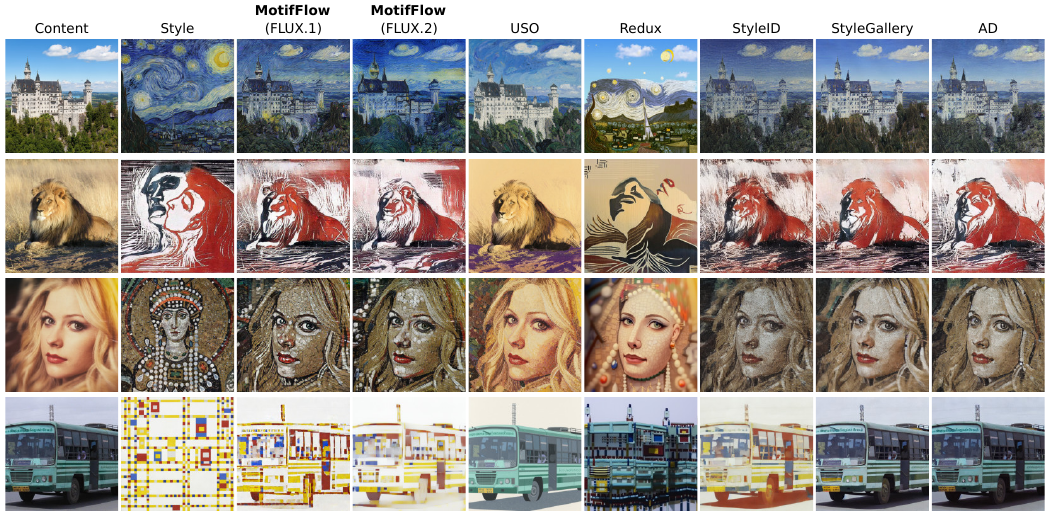}
\caption{\textbf{Style transfer across visual forms.} \method{} adapts painted strokes, graphic marks, mosaic tiles, and geometric patterns to the content composition on FLUX.1 and FLUX.2. Redux uses strength $0.65$. AD denotes the official SD\,1.5 implementation. See Appendix~E for additional comparisons across diverse content images and styles.}
\label{fig:qual}
\end{figure*}

\Cref{fig:qual} illustrates how \method{} adapts distinctive style elements to the content image. Swirling painted marks spread across the castle scene, while mosaic tiles follow the face and hair in the portrait. The bus acquires the style image's rectangular color patterns while retaining its recognizable shape and windows. These results show that style can transform the surfaces and marks within a scene while preserving its composition.

On FLUX.1, the quantitative comparison in \cref{tab:main} shows that \method{} transfers the palette effectively and achieves the lowest texture distance among the compared methods. It also produces stronger stylistic resemblance than USO and Redux. Redux has nearly the same LPIPS as \method{}, but lower CSD and greater structural distance. These results thus demonstrate that our self-contained approach achieves strong style resemblance and local texture fidelity while maintaining recognizable content structure.

\begin{table*}[t]
\centering\small
\setlength{\tabcolsep}{4pt}
\caption{\textbf{Comparison with existing methods and transfer to FLUX.2.} LPIPS and Structure measure changes from the content image. CSD, Color, and Texture measure resemblance to the style image. Failure denotes the style-transfer failure rate (CSD $<0.30$). Metric definitions are given in \cref{sec:exp:setup}. Best values are bold.}
\label{tab:main}
\scalebox{0.95}{%
\begin{tabular}{@{}llcrrrrrr@{}}
\toprule
Method & Backbone & Training-free & LPIPS $\downarrow$ & Structure $\downarrow$ & CSD $\uparrow$ & Color $\downarrow$ & Texture $\downarrow$ & Failure (\%) $\downarrow$\\
\midrule
StyleID~\cite{chung2024styleid} & SD\,1.5 & \checkmark & 0.518 & 0.090 & 0.447 & 6.44 & 0.417 & 11.9\\
StyleGallery~\cite{he2026stylegallery} & SD\,1.5 & \checkmark & 0.463 & 0.094 & 0.515 & 10.40 & 0.200 & 22.5\\
AD~\cite{zhou2025attentiondistillation} & SD\,1.5 & \checkmark & 0.535 & 0.110 & 0.622 & 7.89 & 0.135 & 12.8\\
\midrule
USO~\cite{wu2025uso} & FLUX.1 & -- & 0.538 & 0.110 & 0.331 & 13.06 & 0.349 & 46.8\\
Redux~\cite{bfl2024redux} & FLUX.1 & -- & 0.553 & 0.124 & 0.386 & 17.80 & 0.602 & 30.8\\
\midrule
Injection only & FLUX.1 & \checkmark & 0.490 & \textbf{0.080} & 0.362 & 5.66 & 0.176 & 29.6\\
\method{} & FLUX.1 & \checkmark & 0.560 & 0.110 & 0.592 & \textbf{4.98} & \textbf{0.066} & 1.2\\
\midrule
Injection only & FLUX.2 & \checkmark & \textbf{0.452} & \textbf{0.080} & 0.329 & 8.81 & 0.226 & 44.6\\
\method{} & FLUX.2 & \checkmark & 0.579 & 0.116 & \textbf{0.631} & 6.46 & 0.094 & \textbf{0.3}\\
\bottomrule
\end{tabular}%
}
\end{table*}

On FLUX.1, \method{} reduces the fraction of outputs below CSD $0.30$ from $29.6\%$ with injection alone to $1.2\%$, compared with $12.8\%$ for AD (rates across thresholds: Appendix~D.3). This reduction supports more consistent stylization across inputs. See Appendix~E for further visual comparisons.

To assess transfer to another backbone, we apply the same injection--guidance design to FLUX.2 klein~\cite{bfl2025flux2}, with native style-image conditioning (\cref{tab:main,fig:qual}). Its improvements over injection alone show that guidance also strengthens stylization on FLUX.2 (settings and comparisons: Appendix~G.2).

\subsection{User Study}
\label{sec:exp:user}

The pairwise user study assesses style fidelity, content preservation, and overall preference alongside the automatic metrics. For each baseline, 50 participants evaluate five content--style pairs, yielding 250 judgments per criterion (see Appendix~D.4 for the full protocol).

Participants prefer \method{} to USO and Redux for style fidelity and overall quality, with overall preference rates of $74.4\%$ and $71.6\%$, respectively. Against StyleID and StyleGallery, participants likewise prefer \method{} for style fidelity and overall quality, despite favoring the baselines for content preservation. Preferences are closely balanced against AD, with $51.6\%$ choosing \method{} overall (\cref{tab:user}).

\begin{table}[t]
\centering\small
\setlength{\tabcolsep}{5pt}
\caption{\textbf{User preference (\%).} Percentage of judgments choosing \method{} over each baseline, with 250 judgments per criterion.}
\label{tab:user}
\begin{tabular}{@{}lccc@{}}
\toprule
\method{} vs. & Style & Content & Overall\\
\midrule
USO & 79.2 & 51.2 & 74.4\\
Redux & 75.6 & 53.6 & 71.6\\
StyleID & 67.2 & 46.0 & 63.6\\
StyleGallery & 62.4 & 44.8 & 58.8\\
AD & 53.6 & 48.4 & 51.6\\
\bottomrule
\end{tabular}
\end{table}

\subsection{Stylization Control}
\label{sec:exp:dial}

\begin{figure}[!t]
\centering
\includegraphics[width=0.95\linewidth]{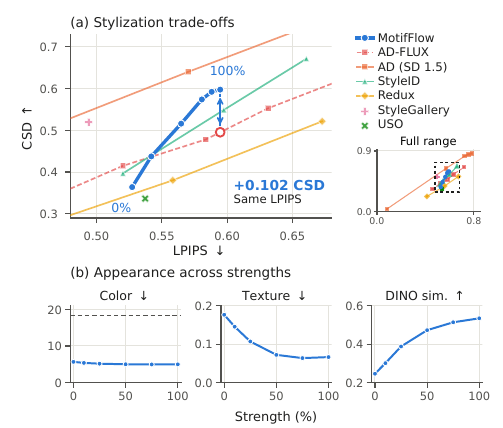}
\caption{\textbf{Strength and content--style trade-offs.} (a) Native control curves, with the full range shown at right. The arrow marks \method{}'s CSD gain over AD-FLUX at the same LPIPS, using linear interpolation. (b) Color and texture distances and DINO image similarity between the output and style image across guidance strengths. The dashed line marks the unmodified content's color distance.}
\label{fig:curves}
\end{figure}

The strength parameter controls how prominently brushwork and motifs appear in the result. \Cref{fig:curves}(b) shows that palette transfer is already strong with injection alone and most of the reduction in texture distance is achieved by half strength. At higher strengths, DINO image similarity continues to rise as the outputs increasingly reproduce the style image's characteristic forms, including the painted strokes, engraved lines, and decorative motifs shown in \cref{fig:teaser}. Intermediate settings therefore offer textured stylization with less prominent motifs, while higher strengths emphasize the style's characteristic forms (see Appendix~F.1 for the complete measurements).

\begin{figure}[!t]
\centering
\includegraphics[width=0.9\linewidth]{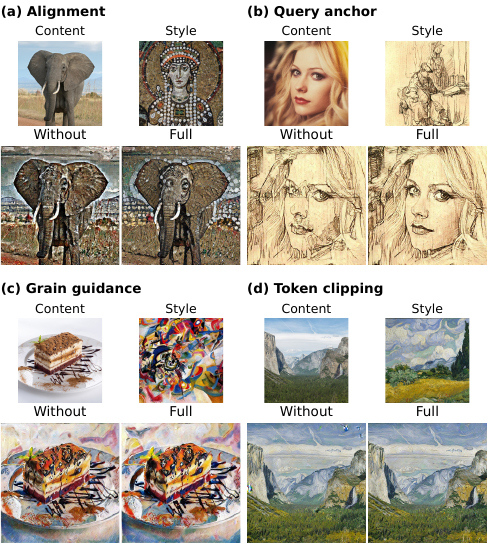}
\caption{\textbf{Qualitative component ablations.} Each panel removes the indicated component. The full method develops coherent mosaic patterns (a), preserves facial structure (b), strengthens color regions and contours (c), and suppresses isolated bright speckles visible in the upper-right sky (d).}
\label{fig:ablation}
\end{figure}

\begin{figure*}[!t]
\centering
\includegraphics[width=0.9\linewidth]{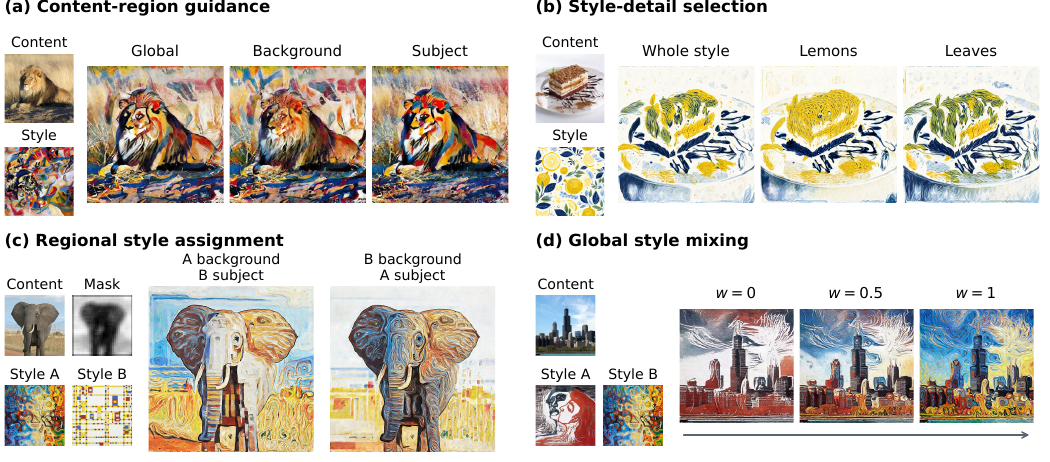}
\caption{\textbf{Spatial and multi-style control.} (a) Global guidance and guidance restricted to the background or subject. (b) Selecting lemon or leaf motifs from one style image. (c) Swapping two styles between background and subject. White in the mask indicates the background. (d) Global mixing from style A ($w=0$) to B ($w=1$).}
\label{fig:spatial}
\end{figure*}

\Cref{fig:curves}(a) compares the style resemblance and perceptual change obtained by varying each method's native control. Our strength sweep increases CSD with a relatively small increase in LPIPS. Redux gains style similarity at the cost of greater perceptual change. To compare coupled sampling with direct latent optimization on FLUX, we port AD's optimization procedure to FLUX.1-dev (\adflux{}) and vary its content-loss weight. At LPIPS $0.595$, \method{} reaches CSD $0.597$, compared with $0.495$ on the interpolated \adflux{} curve, while reducing texture distance by more than half (complete sweeps and interpolation: Appendix~F.2). \method{} thus achieves stronger style resemblance and greater local texture similarity at a comparable level of perceptual change from the content image. \method{} also runs $8.2\times$ faster than \adflux{} on an H100, including both content and style inversion (Appendix~G.1).

\paragraph{Spatial and Multi-Style Control.}
Beyond overall strength, \method{} lets users choose where style develops and which elements it contains. In \cref{fig:spatial}(a,b), regional guidance concentrates additional painted marks on the lion or its background, while style selection changes the cake's decoration from lemon motifs to leaf motifs. \Cref{fig:spatial}(c,d) extends this control to two styles, whose appearances can be assigned to different regions or blended globally (see Appendix~C for mask extraction details and sampling settings).

\subsection{Ablation Study}
\label{sec:exp:ablation}

We remove each component from the full method to examine its effect on style detail, content structure, and local image quality (\cref{tab:ablation,fig:ablation}).

\begin{table}[t]
\centering\small
\setlength{\tabcolsep}{4pt}
\caption{\textbf{Component ablations.} Each $-$ row removes one component from the full method.}
\label{tab:ablation}
\begin{tabular}{@{}lrrrr@{}}
\toprule
Variant & LPIPS $\downarrow$ & Structure $\downarrow$ & CSD $\uparrow$ & Texture $\downarrow$\\
\midrule
Full & .560 & .110 & .592 & .066\\
\midrule
$-$ Alignment & .536 & .091 & .388 & .129\\
$-$ Query anchor & .639 & .134 & .703 & .052\\
$-$ Grain & .542 & .107 & .572 & .069\\
$-$ Clipping & .569 & .114 & .578 & .069\\
\bottomrule
\end{tabular}
\end{table}

\paragraph{Style Alignment.}
Alignment guidance supplies the largest increase in style similarity among the components. Removing it breaks the coherent tile patterns in the mosaic example and nearly doubles texture distance, from $0.066$ to $0.129$. The attention-output constraint therefore helps the image develop the style's characteristic patterns.

\paragraph{Content-Query Anchor.}
The query anchor protects the content structure as style develops. Without it, the portrait's facial features become distorted, and structure distance rises from $0.110$ to $0.134$. CSD increases at the same time, showing the trade-off: stronger resemblance to the style image can come with unwanted changes to the subject. The anchor maintains the facial arrangement while allowing the line-drawing appearance to emerge.

\paragraph{Grain Guidance.}
Grain guidance strengthens local color regions and contours. In the cake example, it develops bolder colored marks and dark outlines that resemble the style painting. This local refinement complements alignment in transformer feature space.

\paragraph{Token-Wise Clipping.}
Clipping suppresses isolated saturated speckles from large local updates. In the landscape example, removing it produces bright spots in the upper-right sky. Clipping removes these spots while retaining the painted texture.

\section{Conclusion}
\label{sec:conclusion}

We introduced \method{}, a training-free, self-contained framework that enables FLUX to transfer distinctive brushwork and motifs while retaining recognizable content structure. Our cross-backbone analysis links the limitations of direct injection to architectural differences in neighboring feature integration. Motivated by this analysis, \method{} couples position-free style injection with latent guidance that aligns native attention outputs with their injected counterparts. This design provides continuous control from palette transfer to pronounced stylization, with regional stylization and style mixing as additional capabilities. Experiments demonstrate more consistent transfer than injection alone and a better quality--efficiency trade-off than direct AD optimization on FLUX. Results on FLUX.2 further demonstrate that expressive style transfer can be achieved using the pretrained backbone's own representations, without auxiliary image encoders or trained adapters.

\makeatletter
\newpage
\if@firstcolumn\else\null\newpage\fi
\makeatother
{
    \small
    \bibliographystyle{ieeenat_fullname}
    \bibliography{references}
}

\clearpage
\appendix
\setcounter{figure}{0}
\setcounter{table}{0}
\renewcommand{\thefigure}{S\arabic{figure}}
\renewcommand{\thetable}{S\arabic{table}}
\renewcommand{\theHfigure}{S\arabic{figure}}
\renewcommand{\theHtable}{S\arabic{table}}
\maketitlesupplementary

This supplement is organized as follows. Each appendix supports the main-paper section given in parentheses. Figures and tables in this supplement carry the prefix S, and numbers without it refer to the main paper.
\begin{itemize}
\item \textbf{\Cref{supp:mechanism}} reports the controlled experiments behind the design choices in the motivation (Sec.~3).
\item \textbf{\Cref{supp:impl}} gives the implementation details of feature caching, latent updates, and grain guidance (Sec.~4).
\item \textbf{\Cref{supp:spatial}} describes the points at which the method can be steered and the regional and multi-style controls that use them (Sec.~4).
\item \textbf{\Cref{supp:eval}} specifies the evaluation images, comparison settings, metrics, and user-study protocol (Sec.~5).
\item \textbf{\Cref{supp:gallery}} provides additional visual comparisons (Sec.~5).
\item \textbf{\Cref{supp:strength}} reports the guidance-strength sweep, the comparison with direct latent optimization, and a comparison of strength controls across methods (Sec.~5).
\item \textbf{\Cref{supp:additional}} reports inference cost and results on FLUX.2 klein (Sec.~5).
\item \textbf{\Cref{supp:limitations}} discusses limitations and future work.
\end{itemize}

\section{Analysis of Attention Injection}
\label{supp:mechanism}

We examine the design choices in the main paper through controlled comparisons. Unless noted otherwise, the diagnostics in this section use the same subset of 200 benchmark pairs, and each sweep varies only the intervention under study. CSD, texture distance, and LPIPS are defined in \cref{supp:eval:metrics}. We first vary RoPE strength at the injection blocks and find that stronger positional encoding pulls the output toward the style image's layout. We therefore omit RoPE at these blocks (\cref{supp:diag:position}). With this position-free injection, the output texture moves closer to the style image, yet switching style images for the same content produces little change in stroke-scale details (\cref{supp:diag:specificity}). We then test whether StyleID's attention sharpening can recover those details. Sharpening beyond standard attention improves U-Net style scores. On FLUX, even the best setting changes the scores only marginally, and stronger sharpening accentuates grid artifacts and lowers style similarity (\cref{supp:diag:sharpening}). To understand this difference, we trace the response to local attention perturbations and compare FLUX attention weights with and without RoPE at fixed queries and keys. Responses are less localized under FLUX injection than in U-Nets, and removing RoPE weakens attention to nearby tokens, helping explain why sharper matches do little to improve coherent brushwork (\cref{supp:diag:spatial}). Finally, we keep injection active and guide the latent to bring the image's own attention output closer to the injected output. Compared with injection alone, this raises texture-based style identification on FLUX.1 from $33\%$ to $88\%$ while largely preserving the content layout (\cref{supp:diag:guidance}).

\subsection{Positional Correspondence and Layout Drift}
\label{supp:diag:position}

We first test whether retaining FLUX's positional encoding during cross-image injection preserves the content composition. We scale image coordinates by $r$ before applying RoPE at injection blocks, while keeping $\tau=1$. A value of one retains the original rotations, and a value of zero removes them. The same setting is used during inversion and generation. Blocks outside the injection set retain their original RoPE. To measure layout, we blur grayscale versions of the output, content, and style images with a Gaussian of standard deviation 16 pixels. We then compute output correlations with the content and style image. These correlations describe coarse luminance structure, not semantic object identity.

\begin{table}[t]
\centering\small
\caption{\textbf{RoPE strength at the injection blocks on FLUX.1.} The layout columns give the correlation between the output's layout and that of the content or style image. As $r$ increases, style similarity rises together with copying of the style image's layout.}
\label{tab:rope}
\begin{tabular}{@{}rrrrrr@{}}
\toprule
$r$ & CSD $\uparrow$ & Texture $\downarrow$ & LPIPS $\downarrow$ & \shortstack{Content\\layout} & \shortstack{Style\\layout}\\
\midrule
0 & .364 & .175 & .527 & .93 & .08\\
1/16 & .375 & .170 & .547 & .91 & .12\\
1/8 & .384 & .166 & .564 & .88 & .13\\
1/4 & .400 & .166 & .598 & .83 & .25\\
1/2 & .451 & .183 & .684 & .70 & .54\\
1 & .602 & .204 & .809 & .51 & .80\\
\bottomrule
\end{tabular}
\end{table}

\begin{figure}[t]
\centering
\includegraphics[width=0.85\linewidth]{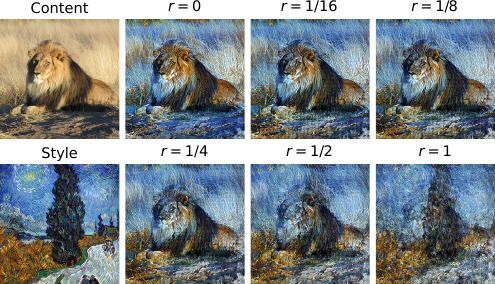}
\caption{\textbf{Injection alone at different RoPE strengths $r$.} The settings are those of \cref{tab:rope}. At $r=1$, the output follows the style image's layout.}
\label{fig:rope}
\end{figure}

Retaining position increases exact adjacency---the fraction of neighboring queries whose strongest style matches preserve their displacement---from $0.54\%$ to $12.8\%$. However, the correspondence is tied to the style image's coordinates. As $r$ rises, correlation with the style image's layout increases from $0.08$ to $0.80$, while content-layout correlation decreases from $0.93$ to $0.51$ (\cref{tab:rope}). \Cref{fig:rope} shows one example, in which the lion remains visible up to $r=1/4$, fades at $r=1/2$, and is replaced by the style image's cypress at $r=1$. The CSD gain accompanies a shift from content structure toward the style image's structure. These measurements explain the choice to remove RoPE at the injection blocks: the style image's appearance can then be introduced without imposing the style image's coordinate system on the content.

We also check whether RoPE should be removed beyond the injection blocks. Compared with removing it only at the injection blocks, removing it from all 57 FLUX.1 blocks lowers CSD from $0.364$ to $0.304$ and raises LPIPS from $0.527$ to $0.540$. We therefore retain RoPE in the other blocks and use position-free injection in the following specificity and sharpening experiments.

\subsection{Style Specificity under Direct Injection}
\label{supp:diag:specificity}

Having limited copying of the style image's layout, we ask whether direct injection transfers the details that distinguish a style. We complement texture distance with style identification and detail correlation. For \emph{style identification}, we match each output to the nearest of the 25 style images under texture distance. Correct identification means that this nearest style image is the one used for generation. Chance accuracy is $4\%$. For \emph{detail correlation}, we compare outputs from the same content under different style images. We subtract grayscale images blurred at standard deviations 2 and 8 pixels, then average their pairwise Pearson correlations. A high value indicates that the location and shape of stroke-scale details change little across style images.

\begin{table}[t]
\centering\small
\caption{\textbf{Style specificity with direct injection.} ID is the rate at which the style image is identified from output texture alone. Detail correlation compares outputs of the same content under different style images. The U-Nets use $\tau=1.5$, and FLUX uses $\tau=1$.}
\label{tab:diag:specificity}
\begin{tabular}{@{}lrrr@{}}
\toprule
Model & Texture $\downarrow$ & ID (\%) $\uparrow$ & \shortstack{Detail\\correlation}\\
\midrule
SD\,1.5 & .381 & 18 & .85\\
SDXL & .391 & 20 & .78\\
FLUX.1 & .175 & 33 & .88\\
FLUX.2 & .280 & 30 & .85\\
\bottomrule
\end{tabular}
\end{table}

\Cref{tab:diag:specificity} shows that limited specificity occurs across the tested architectures. FLUX.1 reduces texture distance from $1.40$ for unmodified content to $0.175$, yet identification remains $33\%$ and detail correlation is $0.88$. Direct injection thus transfers texture without fully adapting the details to each style image. The low identification rates across backbones show that texture similarity and style specificity capture different aspects of stylization. For the FLUX adaptation, the remaining question is whether StyleID's sharpening operation can strengthen the missing detail.

\subsection{Response to Attention Sharpening}
\label{supp:diag:sharpening}

\begin{figure*}[t]
\centering
\includegraphics[width=0.9\linewidth]{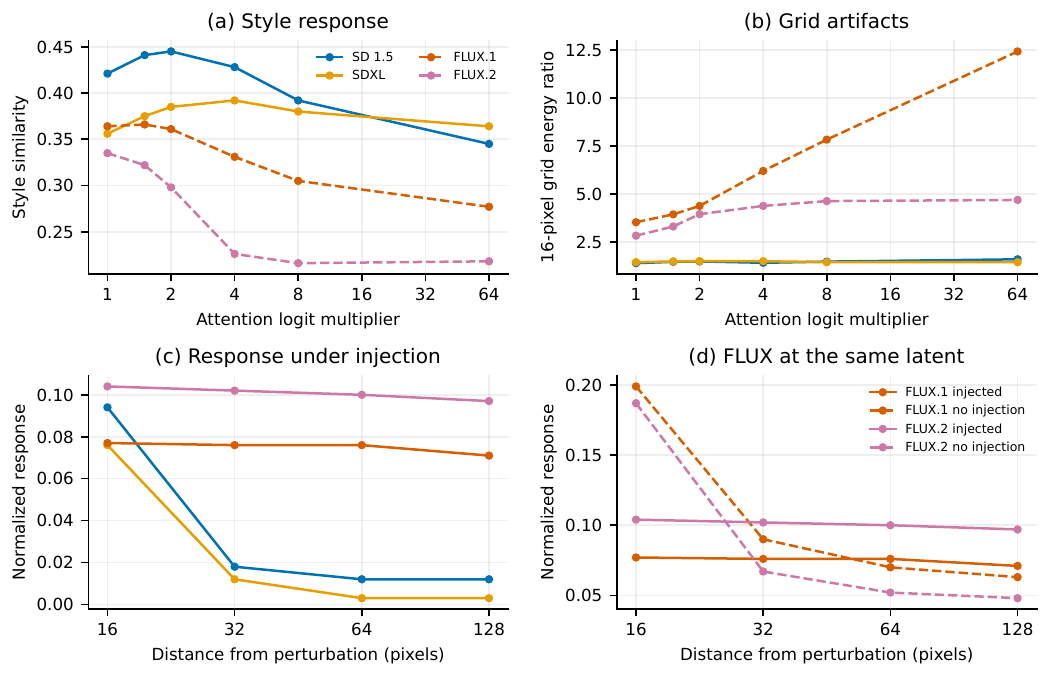}
\caption{\textbf{Attention sharpening and spatial response.} (a,b) Within-model sweeps show that moderate sharpening improves U-Net style scores. On FLUX, sharpening yields almost no gain in style similarity before reducing it, and grid energy increases throughout. (c) At the probed layers, U-Net responses to attention perturbations decay with distance, while FLUX responses under injection vary little. (d) At the same latent, disabling injection across the FLUX transformer restores a stronger response near the perturbed positions. The response is measured at the network output of the current denoising step.}
\label{fig:diag:evidence}
\end{figure*}

To test whether more selective style matching resolves the remaining style deficit, we vary the attention logit multiplier $\tau$ within each model while holding its other settings fixed. Larger values concentrate style attention, and $\tau=1$ is the standard attention of each model. \Cref{fig:diag:evidence}(a,b) plots these sweeps. We report style similarity for each backbone and refer to the texture and specificity measures of \cref{supp:diag:specificity} where they clarify the comparison. For FLUX, we additionally quantify the observed grid artifacts using a grid-energy ratio. For a grayscale image of width $W$, we subtract its mean and compute the squared Fourier magnitude. A 16-pixel period corresponds to frequency bin $k=W/16$. On each horizontal and vertical axis, we divide the energy at $k$ by the median energy at bins $k-6$ through $k-2$ and $k+2$ through $k+6$. The reported ratio averages the two axes. It measures a spectral peak at the token-grid period, rather than general image quality.

On the U-Nets, sharpening beyond standard attention improves the style scores. From $\tau=1$ to the best setting of each model, CSD rises by $0.024$ on SD\,1.5 and $0.036$ on SDXL, and texture distance falls from $0.601$ to $0.334$ and from $0.449$ to $0.354$. The best settings are $\tau=2$ on SD\,1.5 and $\tau=4$ on SDXL, and CSD declines beyond them. FLUX.1 leaves much less room. Its best setting is $\tau=1.5$, where CSD changes only from $0.364$ to $0.366$ and texture distance from $0.175$ to $0.157$. Style identification rises from $33\%$ to $36\%$, and detail correlation stays at $0.88$. At $\tau=2$, CSD, texture distance, and style identification are all worse than at $\tau=1$. Further sharpening increases its grid ratio from $3.52$ at $\tau=1$ to $6.19$ at $\tau=4$, while CSD falls from $0.364$ to $0.331$. On FLUX.2, CSD falls from $0.335$ to $0.322$ already at $\tau=1.5$.

Style scores eventually decline on every backbone once attention becomes too sharp, so the decline itself does not distinguish the architectures. The difference lies in what sharpening offers before that point. U-Nets offer a useful sharpening range with improved style scores, and StyleID's default $\tau=1.5$ lies within it. FLUX has no comparable range. Its best setting leaves the scores nearly unchanged, and the specificity deficit measured in \cref{supp:diag:specificity} remains. Adopting $\tau=1.5$ would therefore not supply the missing detail. Since this setting already raises the FLUX.1 grid ratio to $3.93$ and lowers CSD on FLUX.2, we keep FLUX at standard attention. Visually, the dominant change under sharper FLUX attention is usually stronger color contrast, and sharpening can introduce grid artifacts rather than develop the style image's characteristic brushwork (Fig.~2). Sharpening changes which style features each position retrieves, whereas coherent brushwork also depends on how the backbone integrates these features across neighboring positions. The contrast between the architectures therefore motivates examining how the retrieved features are organized spatially by the backbone.

\subsection{Spatial Processing of Injected Features}
\label{supp:diag:spatial}

\paragraph{Propagation of an injected perturbation.}
We test how a localized change to attention propagates through the remaining network. On 40 matched content--style pairs, we perturb the attention outputs at four well-separated image positions. Each receives a random vector whose RMS matches that of the layer output. We recompute the denoiser output at the same latent and timestep and measure the mean squared change across channels. We group these changes by Chebyshev distance to the nearest perturbed position, divide each profile by its value at the perturbed positions, and average the normalized profiles over pairs and over early, middle, and late denoising steps for each model. The resulting profile describes how far a local change spreads within one denoising forward pass, with distances expressed in image pixels. \Cref{fig:diag:evidence}(c) shows the profiles at SD\,1.5 decoder layer 8, the first attention layer in SDXL module 4, FLUX.1 single-stream block 25, and FLUX.2 single-stream block 16. On SD\,1.5, the normalized response falls from $0.094$ at 16 pixels to $0.012$ at 128 pixels, whereas the FLUX.1 response changes only from $0.077$ to $0.071$.

To measure the effect of injection within FLUX, we repeat the probe at the same latent with injection disabled, so that every block uses its native attention with RoPE. The FLUX.1 responses at 16 and 128 pixels then become $0.199$ and $0.063$ (\cref{fig:diag:evidence}(d)). FLUX.2 changes in the same way, from $0.104$ and $0.097$ under injection to $0.187$ and $0.048$ without it. The broad response is therefore associated with the injection configuration, rather than being an invariant response of FLUX to the same latent. The comparison below, which holds queries and keys fixed, isolates the effect of removing positional rotations on attention weights.

\paragraph{RoPE and local attention.}
The architecture suggests a source of the spatial-processing difference. U-Net convolutions mix neighboring features after attention independently of the style matches. In the FLUX transformer, token mixing is performed by attention, whose positional bias is supplied by RoPE. To measure this contribution directly, we take native queries and keys from eight content images and compute attention with and without their positional rotations, holding the underlying features fixed. We renormalize attention over image keys and measure the mass in a $3\times3$ neighborhood around each query, including the query location itself. We average over heads, queries, the selected injection blocks, and five inversion steps.

\begin{table}[t]
\centering\small
\caption{\textbf{Effect of RoPE on local attention.} Native queries and keys are held fixed. Values are the percentage of image-attention mass within a $3\times3$ neighborhood, averaged over eight content images.}
\label{tab:diag:local}
\begin{tabular}{@{}lrr@{}}
\toprule
Model & With RoPE & Without RoPE\\
\midrule
FLUX.1 & 18.9 & 4.6\\
FLUX.2 & 25.3 & 6.0\\
\bottomrule
\end{tabular}
\end{table}

Removing the rotations substantially reduces local attention (\cref{tab:diag:local}). Together with the perturbation responses, this supports the spatial-processing explanation: U-Nets retain local convolutional mixing when style features are injected, whereas the position-free FLUX adaptation weakens an attention bias toward nearby positions. The layout experiment in \cref{supp:diag:position} explains why retaining that bias during cross-image injection also imports coordinates from the style image.

\paragraph{Style matching and feature averaging.}
We additionally characterize style attention to distinguish direct transport of style neighborhoods from spatial organization by the backbone. For each query, we select the maximum-weight style token. \emph{Exact adjacency} is the fraction of horizontal or vertical query neighbors whose selected style tokens have the same displacement. It measures direct transport of the style neighborhood.

\begin{table}[t]
\centering\small
\caption{\textbf{Feature averaging and exact adjacency in style attention.} $\rho$ is the attention-output variance ratio. The same-image control uses a blurred copy of the style image as content.}
\label{tab:diag:matching}
\begin{tabular}{@{}lrrr@{}}
\toprule
Model & $\rho$ (\%) & \shortstack{Adjacency\\(\%)} & \shortstack{Same-image\\adjacency (\%)}\\
\midrule
SD\,1.5 & 12.0 & 2.97 & 9.0\\
SDXL & 16.2 & 3.44 & 16.4\\
FLUX.1 & 3.6 & .54 & 1.2\\
FLUX.2 & 3.4 & .72 & 3.4\\
\bottomrule
\end{tabular}
\end{table}

The low adjacency rates in \cref{tab:diag:matching} show that direct transport of style neighborhoods through maximum-weight matches is uncommon in both architectures. Even in the same-image control, exact adjacency stays between $1.2\%$ and $16.4\%$. Consequently, understanding coherent output structure requires considering soft attention and subsequent network processing alongside the neighborhoods of the strongest style matches.

We also measure how much feature variation survives attention averaging. For image-only attention $A$, define $X_i=\sum_j A_{ij}V_j$, average access weights $r_j=\frac{1}{n_q}\sum_i A_{ij}$, and $\mu=\sum_jr_jV_j$. The attention-output variance ratio is
\begin{equation}
\rho=\frac{\frac{1}{n_q}\sum_i\|X_i-\mu\|^2}{\sum_jr_j\|V_j-\mu\|^2}.
\end{equation}
The denominator equals the variance across attention outputs plus the mean variance of values under each query's attention weights. Thus $\rho$ describes how much of the variation among style values survives attention averaging, not the fraction of recoverable brushstroke information. For this measurement, attention is renormalized over style image keys, whereas generation includes text keys as well. In \cref{tab:diag:matching}, $\rho$ is $12.0\%$ and $16.2\%$ on the U-Nets and about $3.5\%$ on both FLUX models, so FLUX injection averages away more of this variation. The spread of attention shows the same difference. We measure each query's effective support, $\exp(-\sum_j A_{ij}\log A_{ij})$, as a fraction of the available tokens and average it over queries. Relative to native attention, injection changes this fraction by factors of $0.8$ and $0.9$ on SD\,1.5 and SDXL, and by factors of $3.9$ and $5.6$ on FLUX.1 and FLUX.2. FLUX injection therefore spreads attention much more widely than its native operation.

\subsection{Effect of Latent Guidance}
\label{supp:diag:guidance}

The main paper retains position-free injection and adds an explicit alignment objective between the image's own attention output and the injected attention output. We now test whether this coupled update develops the style-specific detail missing from injection alone. With guidance, style identification rises from $33\%$ to $88\%$, and cross-style detail correlation falls from $0.88$ to $0.64$. Texture distance falls from $0.175$ to $0.066$. The output retains a content-layout correlation of $0.87$, compared with $0.93$ for injection alone, while the grid-energy ratio stays near $3.5$. These results show stronger style-specific detail with a largely preserved content layout and stable grid energy. The component ablations in the main paper identify alignment guidance as the largest contributor to the style gain.

\paragraph{Comparison with partially retained RoPE.}
\Cref{supp:diag:position} showed that retaining part of RoPE at the injection blocks also raises style similarity, at the cost of copying the style image's layout. We therefore compare the two options at the same perceptual change. With RoPE retained at strength $r=1/4$ (\cref{tab:rope}), LPIPS is $0.598$, close to $0.599$ with guidance. At this LPIPS, guidance achieves CSD $0.599$ and texture distance $0.066$, compared with $0.400$ and $0.166$ for partially retained RoPE. Latent guidance thus develops style detail without making the injection matches depend on the style image's coordinate system.

\paragraph{Sensitivity to style detail.}
A perturbation of the style image provides a complementary check that guidance responds to the style image itself. We shuffle the style image's 16-pixel patches, which changes local texture, boundaries, and spatial arrangement at once. As a summary of stroke-scale detail, we take the mean squared difference between Gaussian blurs of the grayscale image at scales 2 and 8 pixels, with intensities in $[0,255]$. Shuffling raises this statistic of the style image from 223 to 615. For FLUX.1 outputs, it changes from 187 to 196 under injection alone and from 237 to 514 with coupled guidance. The larger response under coupled guidance shows greater sensitivity to these changes in the style image.

FLUX.2 follows FLUX.1 in these analyses. It shows the same sharpening and spatial-response trends, and guidance raises its style identification from $30\%$ to $63\%$. The gains on both backbones, together with the component ablations, support alignment guidance as an effective complement to injection: the style image supplies features along the forward trajectory and guides the latent to express style-specific detail through the image's own features.

\section{Implementation Details}
\label{supp:impl}

The method uses style attention in both the forward pass and the latent objective. Reproducing this coupling requires consistent feature caches and noise levels. We specify these choices below.

\subsection{Inversion and Feature Caching}

FLUX.1-dev has 19 double-stream blocks followed by 38 single-stream blocks. We inject at double-stream indices 12--18 and single-stream indices 19--37, zero-indexed within each group. RoPE is omitted in these 26 blocks during both inversions, generation, and gradient evaluation. The other 31 blocks are unchanged. At $512\times512$, the packed image latent contains $32\times32=1024$ tokens, each corresponding to a $16\times16$ image region.

We use an empty text prompt for inversion and generation. The content and style images are encoded by the native VAE. Both are inverted using the reversed 20-step Euler schedule. When moving from $\sigma_{i+1}$ to $\sigma_i$, inversion evaluates the velocity at the destination noise level $\sigma_i$. We cache the content queries and style keys/values from that call for sampling step $i$. The content noise is channel-wise normalized and rescaled to the style noise's mean and standard deviation before generation. Style inversion and its feature bank can be reused for multiple content images.

After each Euler step, style alignment guidance evaluates the transformer once with gradients enabled at the new noise level. The feature-bank index advances to the next sampling index, capped at the final cache entry. Consequently, the last gradient update runs at $\sigma=0$ with the last cached features.

\subsection{Latent Updates and Gradient Normalization}

For the alignment update, we form the clean latent estimate $x_0=x-\sigma v$, where $x$ is the sampling latent after the Euler step and $v$ is the predicted velocity. We evaluate the frozen transformer at $x_0+\sigma v$ and differentiate the guidance loss with respect to $x_0$, holding $v$ fixed. We apply the resulting update to $x$.

The alignment loss compares the two attention outputs defined in the main paper: the injected output $H^{\inj}$, which continues the forward pass, and the auxiliary output $H^{\self}$, which is computed only to measure the loss. For $H^{\self}$, the live image queries attend to the live image and text keys and values, as in ordinary attention. For $H^{\inj}$, the keys and values are the cached style features together with the live text features, and the query mixes the cached content query with the live query at weight $\gamma=0.5$. Text-token outputs keep ordinary attention, and both outputs use the standard attention scale without extra sharpening. The loss is the mean squared difference between $H^{\self}$ and $H^{\inj}$ over image queries, heads, and channels, summed over the injection blocks. The query anchor adds the mean absolute difference between the live and cached content queries, summed over the same blocks, with weight $0.3$.

For a gradient $g$, let $r$ be its global RMS and $r_i$ its RMS at token $i$. We cap the token gradient using
\begin{equation}
\widetilde g_i=g_i\min\left(1,\frac{2r}{r_i+\epsilon}\right),
\end{equation}
then divide by the original global RMS:
\begin{equation}
\bar g_i=\frac{\widetilde g_i}{\max(r,10^{-12})}.
\end{equation}
The result is the processed gradient $\bar g$ used in the main-paper update. Using the original RMS limits isolated updates without renormalizing them back upward after clipping. A content-region mask, when used, multiplies the clipped gradient before the latent update.

\begin{table}[t]
\centering\small
\caption{\textbf{Default FLUX.1 settings.} The same settings are used for all benchmark pairs.}
\begin{tabular}{@{}lr@{}}
\toprule
Parameter & Value\\\midrule
Resolution & $512\times512$\\
Sampling steps & 20\\
Content-query blend $\gamma$ & 0.5\\
Attention logit multiplier & 1\\
Alignment step, before $28/N$ scaling & 0.05\\
Grain step, before $28/N$ scaling & 0.025\\
Query-anchor weight & 0.3\\
Alignment noise interval & $[0,1]$\\
Grain noise interval & $[0.15,0.85]$\\
Token RMS cap / global RMS & 2\\
\bottomrule
\end{tabular}
\end{table}

Both nominal step sizes are multiplied by $28/N$ for $N$ sampling steps, with $N=20$ on FLUX.1, and then by the user-selected strength $s$. Alignment guidance is applied at every step, including the final update at $\sigma=0$, whereas grain guidance is active only for $\sigma\in[0.15,0.85]$.

\subsection{Local Texture Statistics}

Grain guidance operates on the clean estimate arranged as a 16-channel latent image. We remove spatial frequencies below 0.10 cycles per latent pixel. At the first scale, each position stacks its $3\times3$ neighborhood into a 144-dimensional vector. At the second scale, the high-pass latent is first average-pooled by a factor of two, then the same neighborhood features are extracted. If $\Phi_k$ (a $144\times n_k$ matrix) contains these vectors, $G_k=\Phi_k\Phi_k^\top/n_k$ is its Gram matrix. For the current clean latent estimate $z$ and style latent $z^s$, the texture loss is
\begin{equation}
\mathcal L_G=\sum_{k=1}^2
\frac{\|G_k(z)-G_k(z^s)\|_F^2}{\|G_k(z^s)\|_F^2+\epsilon}.
\label{eq:gram}
\end{equation}
The denominator normalizes each discrepancy by the style Gram matrix's squared Frobenius norm, with $\epsilon$ protecting against zero denominators.

The normalized, token-clipped gradient acts directly on the clean estimate, and its displacement is added to the sampling latent. It does not differentiate through the velocity prediction. This makes the refinement inexpensive and complementary to alignment guidance, whose gradient does traverse the transformer. The strength control scales both updates, including the query anchor within the alignment objective.

\section{Control Points for Regional and Multi-Style Stylization}
\label{supp:spatial}

\begin{figure*}[t]
\centering
\includegraphics[width=\linewidth]{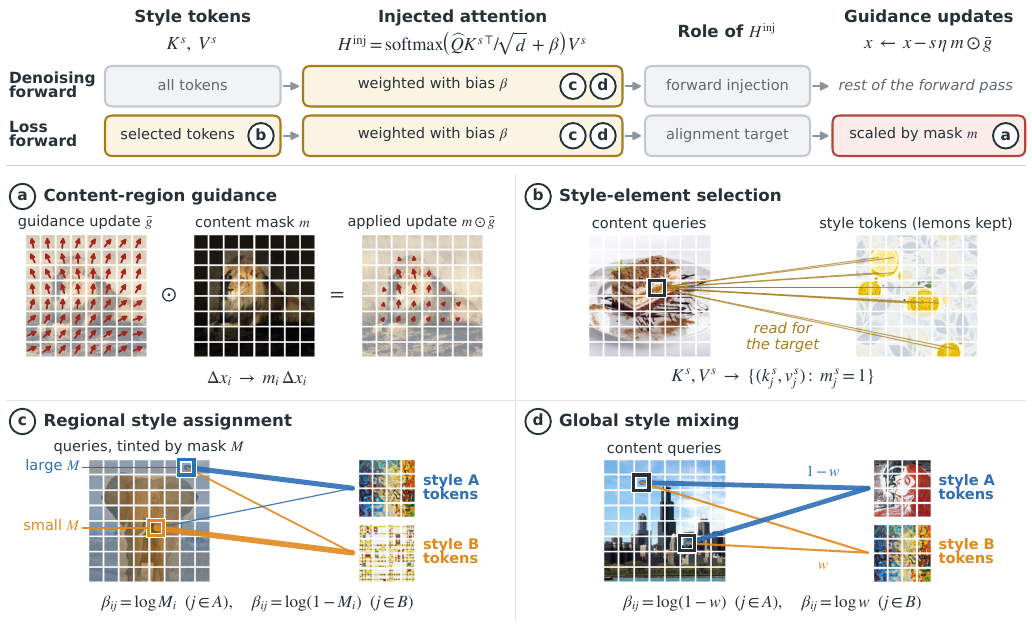}
\caption{\textbf{Three points at which a sampling step can be steered, shown through the controls of Fig.~7.} Top: the two forward passes of one sampling step, with text tokens omitted. Both passes read the style tokens through injected attention. In the denoising forward, the output $H^{\mathrm{inj}}$ replaces the block's attention output, which is forward injection. In the loss forward, $H^{\mathrm{inj}}$ serves as the alignment target, and the resulting gradient updates the latent. The colored cells are the three points: the style tokens that are read, the bias on the attention logits, and the mask on the guidance update. Letters mark the control of Fig.~7 that uses each point. (a) The guidance update is multiplied token by token with a soft mask of the content image, so that the latent changes only inside the chosen region. (b) When the alignment target is computed, each query reads only the style tokens of the selected element. (c) With two styles, a bias $\beta$ on the attention logits steers each query toward style A where the mask $M$ is high and toward style B where it is low. (d) Global mixing uses the same bias with constant weights $1-w$ and $w$ at every query. The token grids are cut from the inputs of the Fig.~7 examples and use the masks of those runs at reduced resolution. Arrow directions and line widths are schematic.}
\label{fig:spatial:mech}
\end{figure*}

\method{} can be steered at three points of a sampling step: which style tokens are read, how the attention over them is weighted, and how the guidance update is applied to the latent. \Cref{fig:spatial:mech} (top) places these points in the two forward passes. The four controls in Fig.~7 each use one of them, and we describe the controls here to show what each point governs (\cref{fig:spatial:mech}a--d). We first cover the two points that take a mask and then the attention weighting, which we use to combine two styles.

\subsection{Masking the Guidance Update and the Style Tokens}

We extract masks from the input images using the backbone's text-to-image attention. Two descriptions distinguish the regions or elements of interest, such as an animal and its background, or lemons and leaves. For each description, we average attention across text tokens, heads, layers, and noise levels, then normalize the spatial map to sum to one. The difference between the two maps identifies locations associated more strongly with the first description.

The guidance update is the point that decides where additional stylization develops. For content-region guidance, the difference map is converted to a smooth mask $m$ of the content image, which multiplies both latent guidance updates token by token (\cref{fig:spatial:mech}a). The set of style tokens that is read decides which style elements guidance encourages. For style-element selection, the difference map is thresholded into a binary mask $m^s$ over the style tokens, and only the keys and values with $m^s_j=1$ are read when the alignment target is computed in the loss forward (\cref{fig:spatial:mech}b). Neither operation changes the denoising forward, so forward injection remains global.

\subsection{Weighting Attention across Two Styles}

The attention weighting decides how much each query reads from each group of style tokens, and we use it to combine two styles. We concatenate the cached keys and values of the two styles, which we call A and B as in Fig.~7. For regional assignment, a content mask $M$ with values in $[0,1]$, extracted from the content image with the two-description procedure above, adds biases $\beta_{ij}=\log M_i$ and $\beta_{ij}=\log(1-M_i)$ to the attention logits between query $i$ and the tokens $j$ of style A and of style B, respectively (\cref{fig:spatial:mech}c). A query with a large mask value therefore reads mainly from style A, and a query with a small value reads mainly from style B. Global mixing uses the same mechanism with spatially constant weights $1-w$ and $w$ (\cref{fig:spatial:mech}d). The biases enter before softmax and apply in both forward passes, so this point changes forward injection as well as the alignment target.

Regional assignment matches the initial content noise statistics to style A. Global mixing instead matches them to $\sqrt{1-w}\,\epsilon^{A}+\sqrt w\,\epsilon^{B}$, where $\epsilon^{A}$ and $\epsilon^{B}$ are the inverted noises of the two styles.

\subsection{Additional Results}

\begin{figure*}[tp]
\centering
\includegraphics[width=0.8\linewidth]{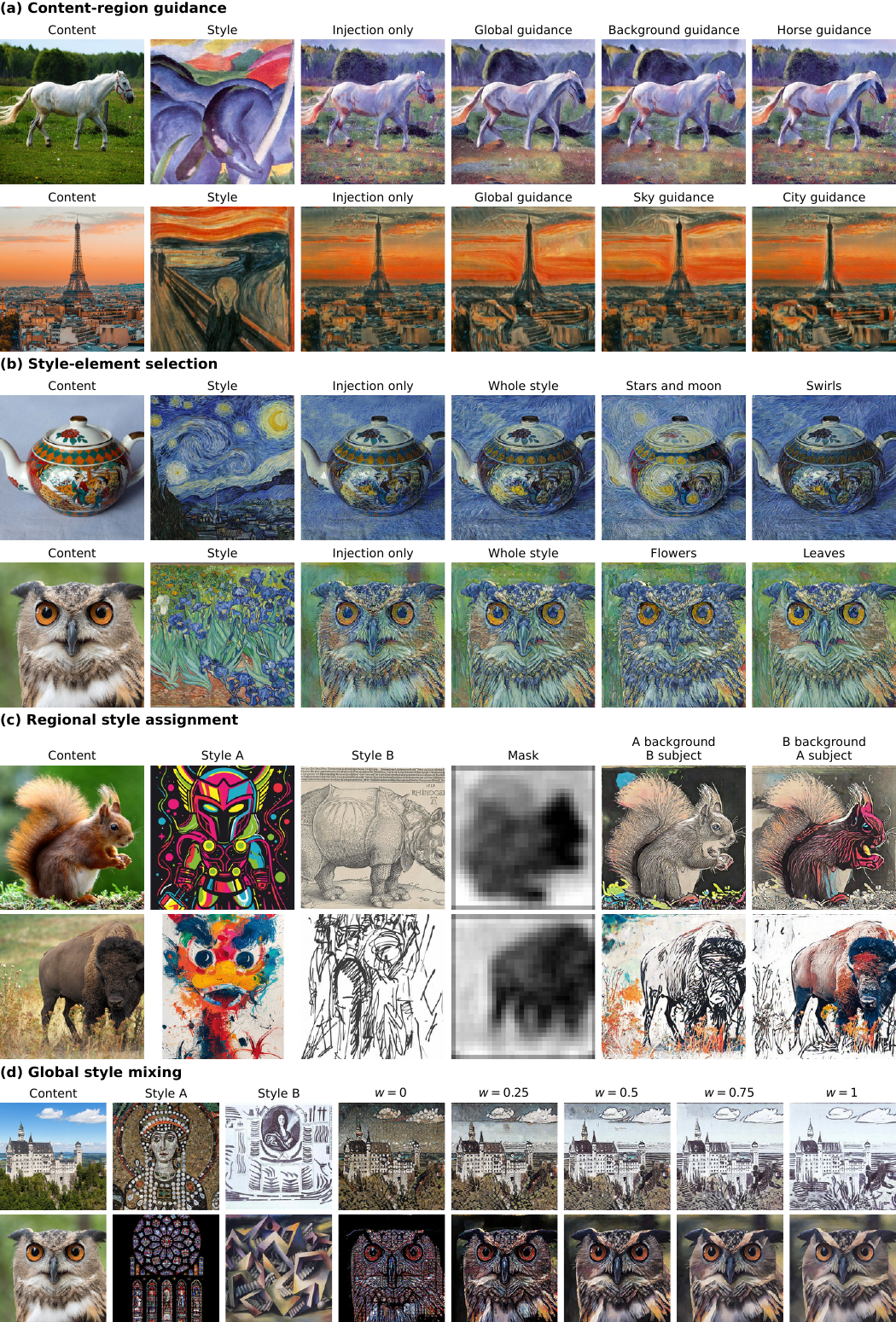}
\caption{\textbf{Additional examples of regional and multi-style control.} The panels follow Fig.~7. (a) Guidance applied globally or restricted to one content region. (b) Guidance toward the whole style image or toward selected elements of it. (c) Two styles swapped between background and subject, where white in the mask marks the background. (d) Global mixing from style A ($w=0$) to style B ($w=1$).}
\label{fig:spatial:extra}
\end{figure*}

\Cref{fig:spatial:extra} gives two further examples of each control in Fig.~7. The mixing examples in (d) include the intermediate weights omitted from Fig.~7.

\section{Evaluation Protocol}
\label{supp:eval}

\subsection{Evaluation Images}
\label{supp:eval:data}

The benchmark combines 40 content images with 25 style images to form 1,000 content--style pairs. Styles span oil painting, ink, watercolor, woodcut, and graphic designs. Content includes people, animals, buildings, landscapes, interiors, and objects. Inputs are center-cropped and resized to $512\times512$.

\subsection{Comparison Settings}
\label{supp:eval:baselines}

\paragraph{Main Comparison.}
For Table~1, we run StyleID~\cite{chung2024styleid}, StyleGallery~\cite{he2026stylegallery}, and AD~\cite{zhou2025attentiondistillation} on SD\,1.5 using their official implementations. StyleID uses its default query-preservation weight $\gamma=0.75$ and attention sharpening $T=1.5$. StyleGallery uses its default content-loss weight of $0.26$. AD uses its default content-loss weight of $0.25$ and 200 optimization steps.

On FLUX, USO~\cite{wu2025uso} uses its official layout-preserving mode with the content and style images, an empty prompt, and guidance scale $4$. It generates at $1024$ resolution, and outputs are resized to $512$ for evaluation. For Redux~\cite{bfl2024redux}, we condition the standard FLUX image-to-image pipeline on the adapter's style-image embeddings and initialize generation from the content image, using guidance scale $2.5$. We use image-to-image strength $0.65$ because its mean LPIPS is closest to that of default \method{} among the tested strengths. Injection alone and \method{} use the settings in \cref{supp:impl}, with guidance strength $s=0$ and $s=1$, respectively.

\paragraph{Control Curves.}
Figure~5 varies each method's native control to trace its content--style trade-off. We vary StyleID's query-preservation weight over $\{0.9,0.75,0.5,0.3\}$, AD's content-loss weight over $\{0.50,0.25,0.15,0.10,0.05\}$, and Redux's image-to-image strength over $\{0.5,0.65,0.8\}$. The \method{} strength settings are listed in \cref{supp:dial}. \Cref{supp:optimization} gives the implementation and content-weight sweep for \adflux{}, our FLUX port of AD.

\subsection{Evaluation Metrics}
\label{supp:eval:metrics}

\paragraph{Color.}
The Color metric uses sliced Wasserstein-2 distance to compare the CIELAB pixel-color distributions of the output and style images. Both images are resized to $512\times512$, and colors are sampled on a stride-4 grid. We project the colors onto 64 fixed random unit directions and sort each projection. The distance is the square root of the mean squared difference between corresponding sorted values, averaged over samples and directions. Unmodified content has a distance of 18.4 on this benchmark. This metric assesses palette transfer independently of where colors are placed.

\paragraph{Texture.}
We compare local texture in grayscale as a proxy for brushwork similarity. To emphasize fine detail, we subtract a Gaussian blur ($\sigma=4$ pixels) from each grayscale image and extract VGG-19 features at \texttt{relu1\_1} and \texttt{relu2\_1} from the normalized residual. The texture distance sums the mean squared errors between the output and style images' normalized Gram matrices at these two layers and scales the result by $10^4$. On this scale, unmodified content images have a mean distance of $1.40$ from their paired style images.

\paragraph{Style Similarity.}
Style similarity is the cosine similarity between the CSD embeddings of the output and style image~\cite{somepalli2024csd}. As a complementary measure, DINO image similarity compares the normalized DINO ViT-B/8 class-token embeddings of the output and style image~\cite{caron2021dino}. In the strength analysis, we interpret its trend alongside the visual examples in Fig.~1 to assess how the outputs adopt the style image's characteristic forms.

\paragraph{Structure.}
The Structure distance is the mean absolute difference between the output and content images' DINO self-similarity matrices, computed from normalized final-layer patch features. This measures changes in spatial feature relations.

\paragraph{LPIPS.}
LPIPS~\cite{zhang2018lpips} measures perceptual difference between the output and content image. It is zero for unchanged content and approximately $0.84$ when the style image itself replaces the content. LPIPS and Structure are read jointly with style metrics: changing appearance can increase LPIPS even when the scene remains recognizable.

\begin{figure}[t]
\centering
\includegraphics[width=0.9\linewidth]{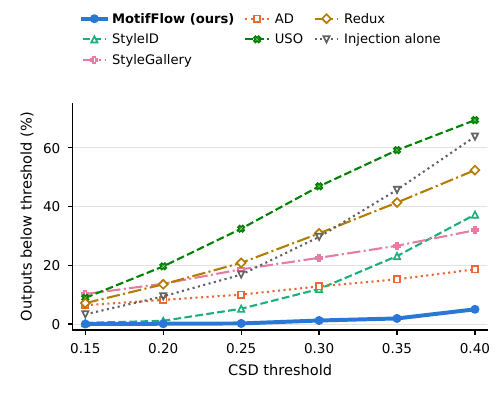}
\caption{\textbf{Low-CSD rates across thresholds.} Each curve gives the percentage of a method's outputs whose CSD falls below the threshold. Lower is better.}
\label{fig:low_csd_rates}
\end{figure}

\paragraph{Low-CSD Rates.}
To assess how consistently each method transfers style, we measure the percentage of outputs below a CSD threshold. \Cref{fig:low_csd_rates} compares this percentage across thresholds from $0.15$ to $0.40$ using the method configurations in Table~1. \method{} has the lowest rate at every tested threshold, supporting more consistent stylization across inputs.

\subsection{User Study Protocol}
\label{supp:userstudy}

The study compares \method{} with USO, Redux, StyleID, StyleGallery, and AD. The same 50 participants each evaluate 25 comparisons, five for each baseline, giving 250 judgments per baseline for each criterion. Each comparison displays the content image, the style image, and the two generated outputs, with their left--right order randomized.

Participants choose one output separately for style fidelity, content preservation, and overall preference. For each baseline and criterion, we divide the number of judgments choosing \method{} by 250 and report the result as a percentage in Table~2. Overall preference is judged directly rather than computed from the style and content responses.

\clearpage
\onecolumn
\section{Additional Visual Comparisons}
\label{supp:gallery}

Each row shows the same content--style pair across methods. The baselines use the main comparison settings in \cref{supp:eval:baselines}. \Cref{fig:gallery:ours} shows further results of \method{} alone.

\begin{center}\includegraphics[width=0.85\linewidth]{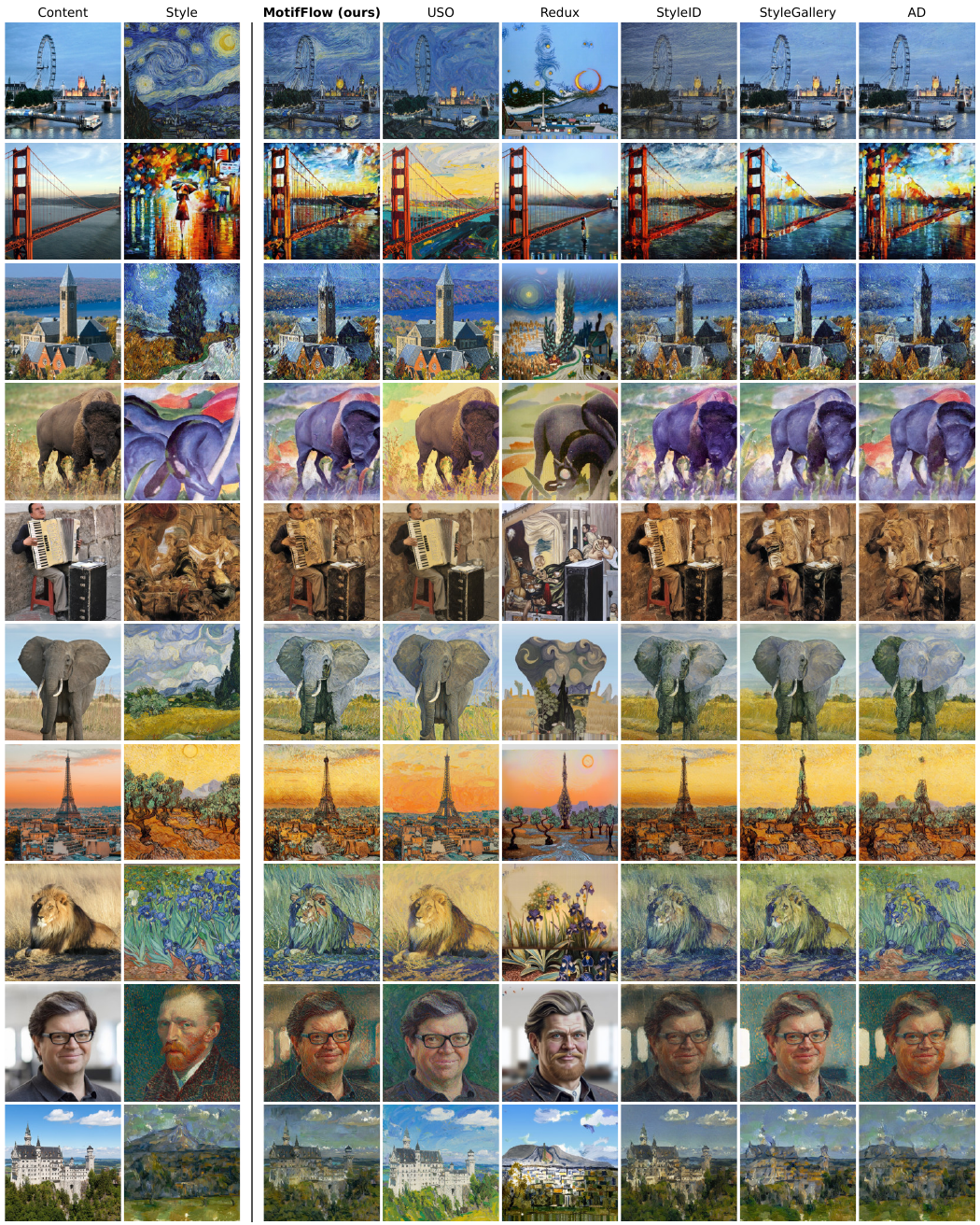}\captionof{figure}{Additional qualitative comparisons: painterly color and brushwork.}\end{center}
\clearpage\begin{center}\includegraphics[width=0.9\linewidth]{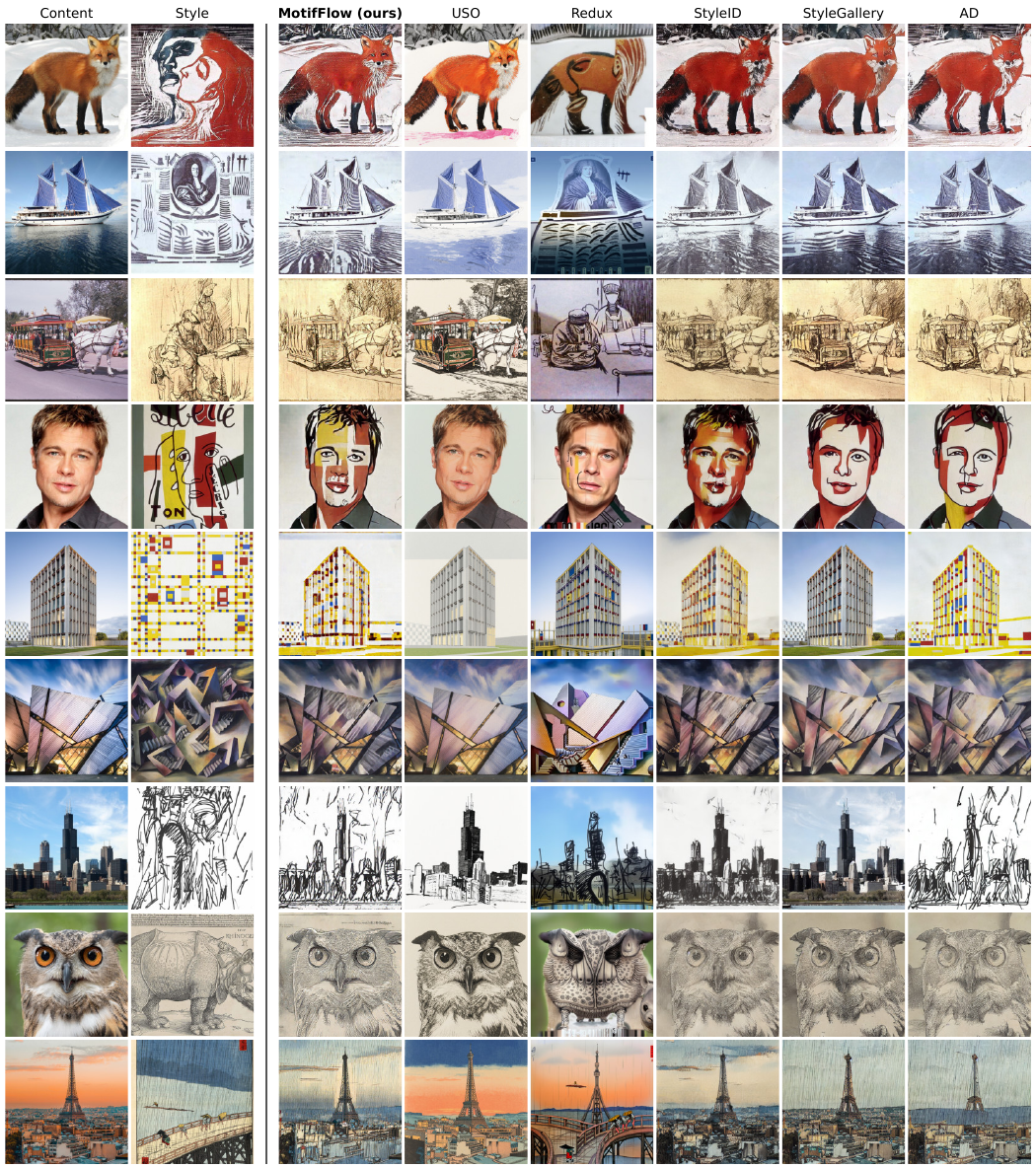}\captionof{figure}{Additional qualitative comparisons: linework and geometric structure.}\end{center}
\clearpage\begin{center}\includegraphics[width=0.9\linewidth]{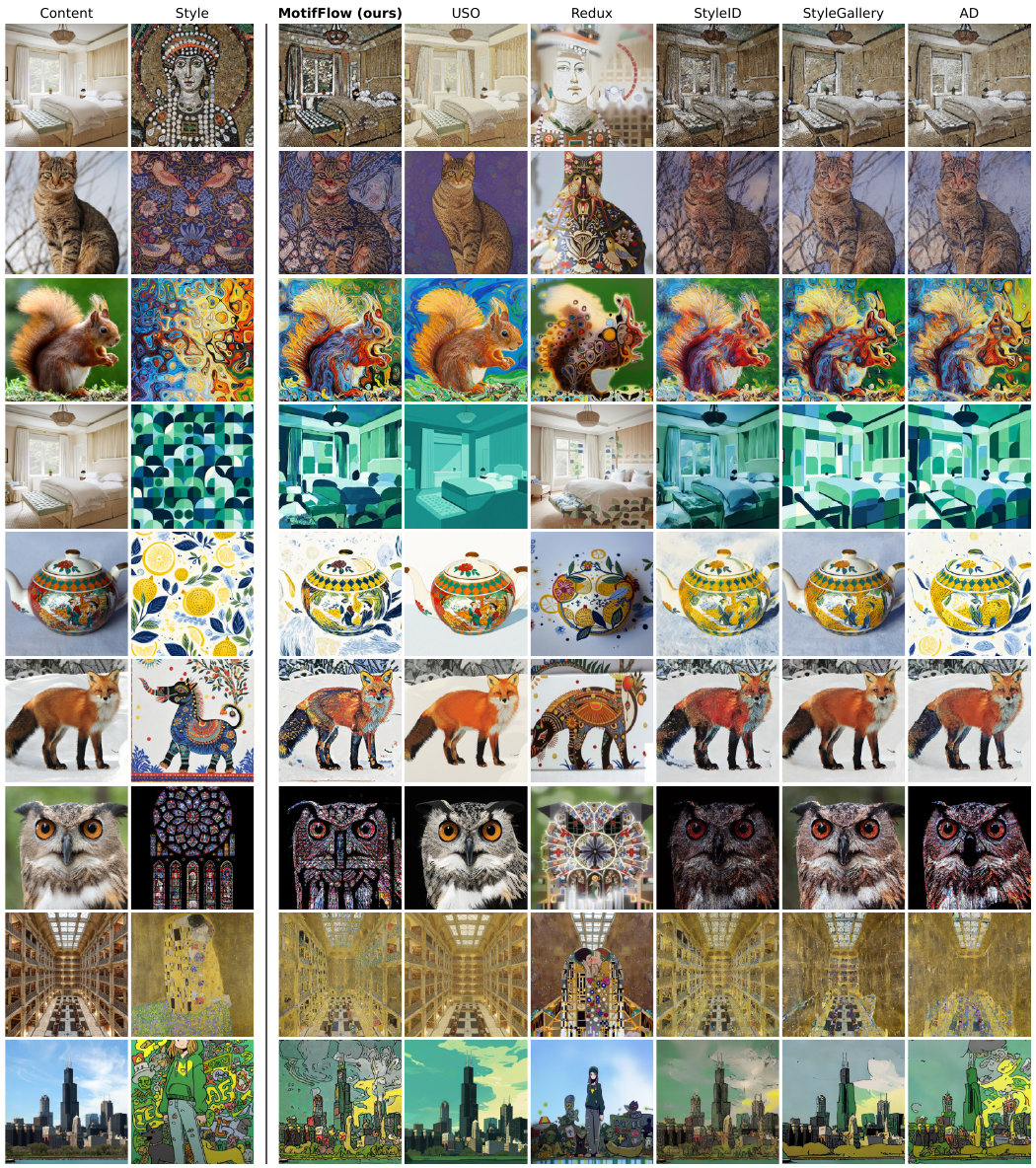}\captionof{figure}{Additional qualitative comparisons: decorative motifs and surface patterns.}\end{center}
\clearpage\begin{center}\includegraphics[width=0.9\linewidth]{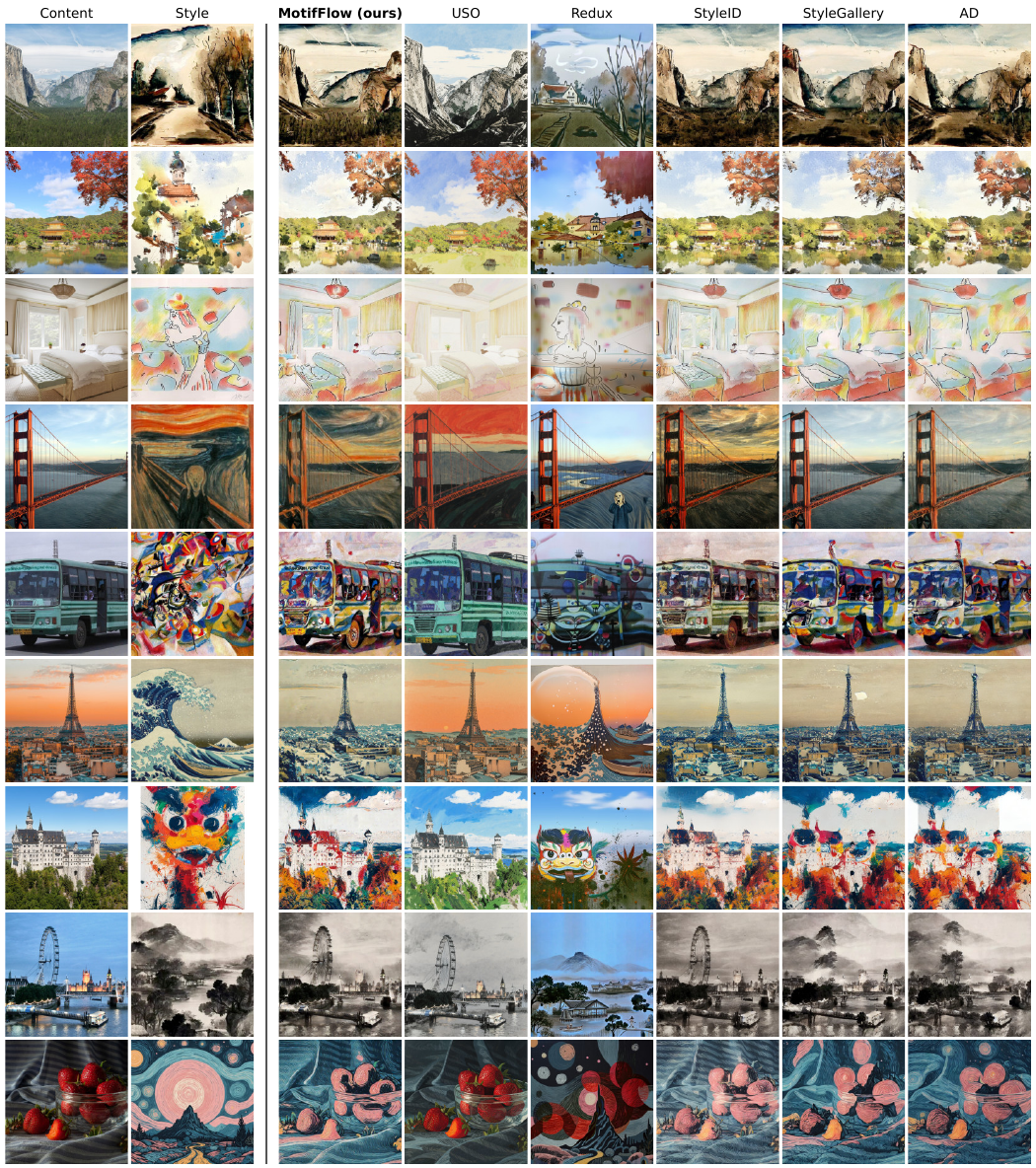}\captionof{figure}{Additional qualitative comparisons: watercolor, expressive color, and wave forms.}\end{center}
\clearpage\begin{center}\includegraphics[width=0.9\linewidth]{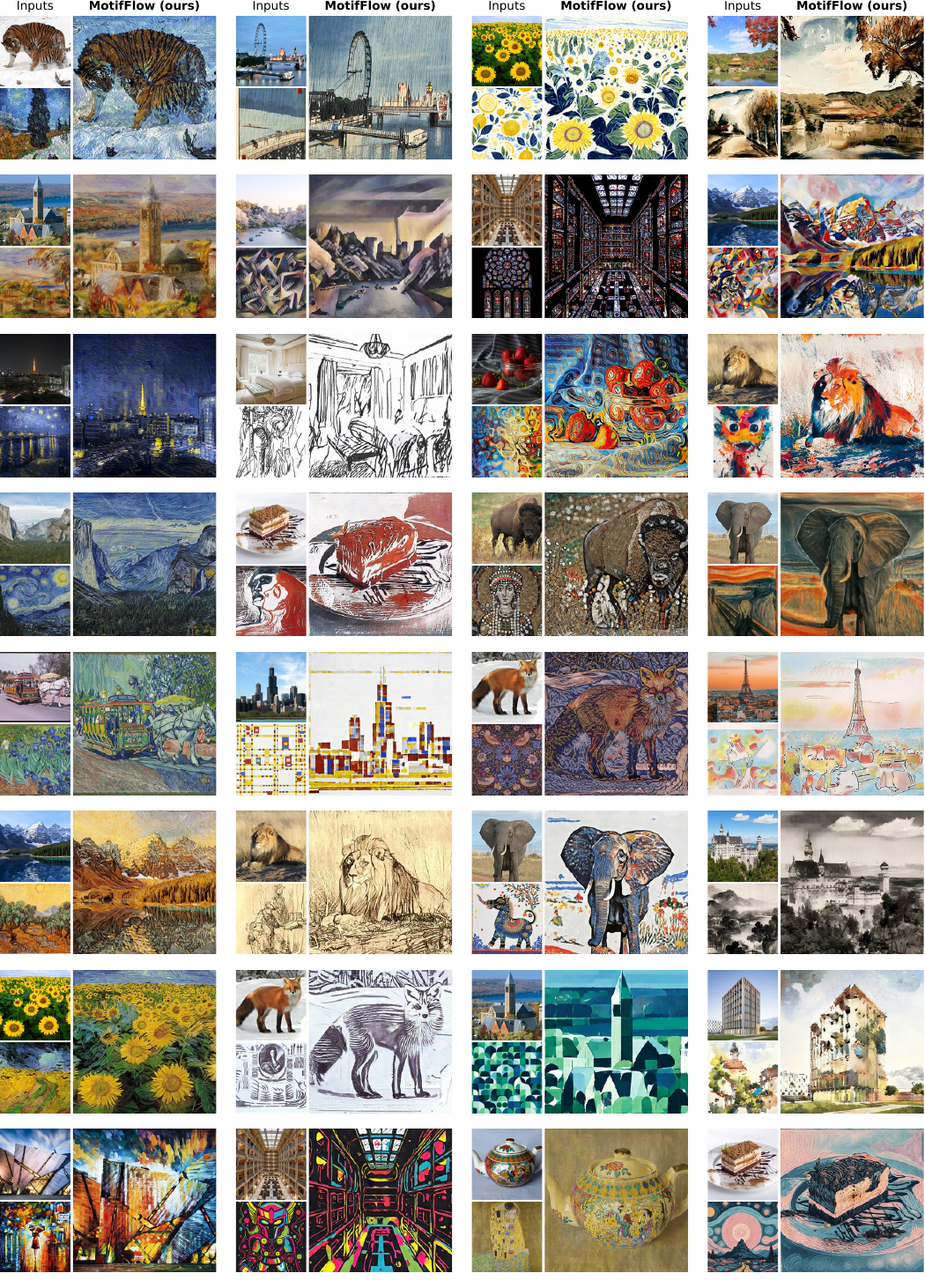}\captionof{figure}{Additional \method{} results. Each group shows the content image (top left), the style image (bottom left), and the result.}\label{fig:gallery:ours}\end{center}
\twocolumn

\section{Stylization Control and Alternative Optimization}
\label{supp:strength}

We first examine how guidance strength changes the output, then compare our coupled sampling procedure with direct latent optimization, and finally compare the strength controls of all methods when one set of settings is applied to every image.

\subsection{Effect of Guidance Strength}
\label{supp:dial}

\begin{table}[t]\centering\small\setlength{\tabcolsep}{3.5pt}
\caption{\textbf{Metrics across guidance strengths.} Strength $0\%$ corresponds to injection alone and $100\%$ to default \method{}.}\label{tab:suppdial}
\begin{tabular}{@{}rrrrrr@{}}\toprule
Strength (\%) & Color $\downarrow$ & LPIPS $\downarrow$ & CSD $\uparrow$ & DINO $\uparrow$ & Texture $\downarrow$\\\midrule
0 & 5.66 & 0.490 & 0.362 & 0.246 & 0.176 \\
10 & 5.38 & 0.504 & 0.433 & 0.301 & 0.145 \\
25 & 5.16 & 0.529 & 0.513 & 0.388 & 0.107 \\
50 & 5.00 & 0.546 & 0.569 & 0.473 & 0.072 \\
75 & 4.96 & 0.554 & 0.587 & 0.514 & 0.064 \\
100 & 4.98 & 0.560 & 0.592 & 0.534 & 0.066 \\\bottomrule\end{tabular}\end{table}

The strength parameter scales the alignment and grain updates, with injection active at every setting. \Cref{tab:suppdial} shows that palette transfer is already strong with injection alone, while guidance substantially improves local texture. Texture distance approaches its full-guidance value by $50\%$. Above this strength, color and texture distances change little, while global DINO similarity continues to rise. The visual examples in Fig.~1 show how these higher strengths develop more pronounced stylistic forms, giving users control beyond the early gains in palette and texture.

\subsection{Comparison with Alternative Optimization}
\label{supp:optimization}

We compare the coupled method with an alternative use of the attention-based objective: direct optimization of a clean latent. The configuration below specifies how this procedure uses the backbone and preserves content.

\paragraph{Direct Optimization on FLUX.}
We port AD's SD\,1.5 style-transfer optimization to FLUX.1-dev and denote it \adflux{}. It updates the clean content latent directly, without forward injection or a denoising trajectory. We keep AD's default optimization settings: 200 Adam updates at learning rate $0.05$. The objective combines an L1 loss between the current attention output and a detached style-attention target with a weighted L1 loss between current and content queries. We follow the official FLUX guidance implementation's layer selection and attention computation, evaluating the losses in single-stream blocks 31--37 over the joint text/image sequence with RoPE retained. For this direct-optimization port, we decrease the transformer's noise-level input linearly toward zero over the updates.

\begin{table}[t]
\centering\small
\caption{\textbf{AD-FLUX content-weight sweep.} Lower weights give stronger style at the cost of greater perceptual change.}
\label{tab:adflux}
\setlength{\tabcolsep}{4pt}
\begin{tabular}{@{}rrrrrr@{}}
\toprule
Weight & LPIPS $\downarrow$ & Color $\downarrow$ & CSD $\uparrow$ & DINO $\uparrow$ & Texture $\downarrow$\\\midrule
.80 & .459 & 7.19 & .331 & .278 & .246\\
.70 & .520 & 5.74 & .415 & .352 & .164\\
.60 & .584 & 4.66 & .478 & .423 & .149\\
.50 & .631 & 3.78 & .553 & .510 & .128\\
.25 & .719 & 2.52 & .658 & .638 & .105\\
\bottomrule
\end{tabular}
\end{table}

The sweep in \cref{tab:adflux} raises the content weight from AD's default of $0.25$ and traces the trade-off between style resemblance and perceptual change. We linearly interpolate between neighboring sweep means to compare the methods at the same LPIPS. At LPIPS $0.595$, \method{} reaches CSD $0.597$ and texture distance $0.065$, compared with $0.495$ and $0.144$ for \adflux{}. Color distance favors \adflux{}, at $4.45$ versus $4.88$. At this level of perceptual change, \method{} achieves stronger overall style resemblance and closer local texture, while \adflux{} matches the palette more closely.

\paragraph{Concurrent Work.}
Hyperion~\cite{li2026hyperion} is a concurrent training-free method for FLUX backbones that also updates the latent during denoising by backpropagating an attention-output matching loss through the frozen transformer. Its authors likewise observe that forward injection alone transfers little texture on these backbones, which agrees with our analysis in \cref{supp:mechanism}. The two methods differ in how they use this objective. Hyperion targets style-guided generation from noise, supplies the content image as additional reference tokens, and applies the optimization without forward injection. \method{} instead starts from the inverted content image, keeps injection active throughout sampling, and uses the output of this injection as the alignment target. As the work is concurrent, we do not compare with it experimentally.

\subsection{Strength Controls across Methods}
\label{supp:controls}

Figure~5 compares the methods through the mean content--style trade-off that each method's native control traces. In practice, however, a user chooses a setting once and expects it to act in the same way on different images. We therefore fix five settings for each method, apply them unchanged to every content--style pair, and examine how the result changes from one setting to the next.

\paragraph{Settings.}
We derive the five settings from the range of each control and from the method's published default, so that they do not depend on the outcome of the comparison. A control defined on $[0,1]$ is sampled over its whole range in equal steps. This gives guidance strengths of $0$, $25$, $50$, $75$, and $100\%$ for \method{}, $\gamma\in\{1,0.75,0.5,0.25,0\}$ for StyleID's query-preservation weight, and $\{0,0.25,0.5,0.75,1\}$ for Redux's image-to-image strength, where strength $0$ returns the content image. A content-loss weight has no upper bound, so we sample it in equal steps from twice the default down to zero, which places the default at the middle setting. The resulting weights are $\{0.5,0.375,0.25,0.125,0\}$ for AD and $\{0.52,0.39,0.26,0.13,0\}$ for StyleGallery. These two ranges span both ends of what the weights can produce. At twice the default, both methods already return the content image, as shown by a mean CSD to the style image of $0.06$, which equals that of the unmodified content images. At zero, the content term is switched off, which is the strongest stylization the weight can give. To show what lies between the five settings, we additionally sweep every control over the same range. \method{} and StyleID are sampled in steps of $5\%$ and $0.05$, which gives 21 values each. The other three controls are sampled more finely where their results change fastest, down to steps of $0.005$ for AD and StyleGallery, which gives 44, 46, and 41 values for AD, StyleGallery, and Redux. USO has no strength control and contributes a single result per pair. All remaining settings follow \cref{supp:eval:baselines}. The comparison covers 50 content--style pairs.

\begin{figure*}[t]
\centering
\includegraphics[width=0.93\linewidth]{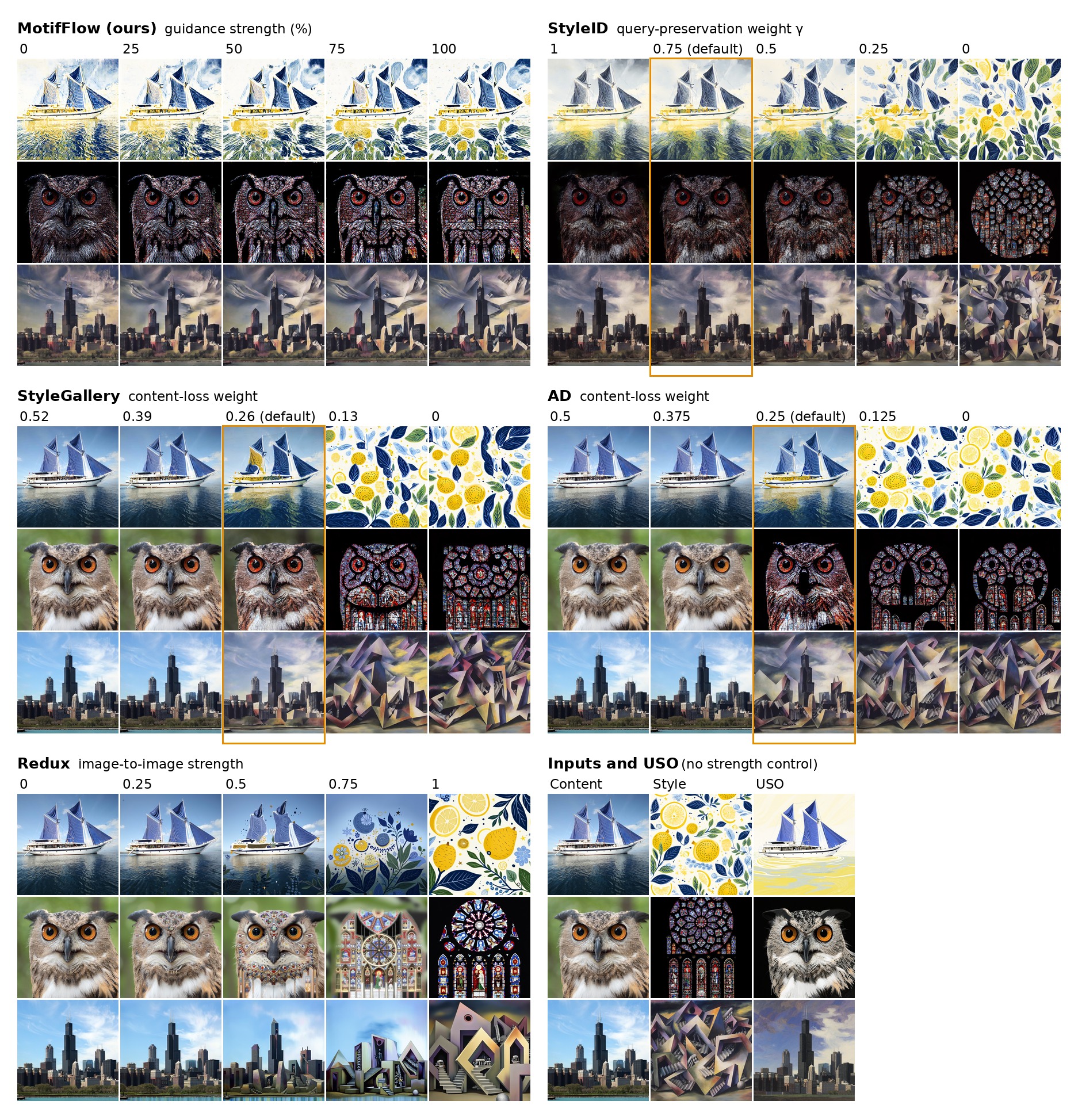}
\caption{\textbf{Five fixed settings of each method's strength control on three content--style pairs.} Each of the first five blocks shows one method. A block's rows are the same three pairs, and its columns are the five settings, ordered from weakest to strongest stylization and shared by all pairs. The last block shows the content and style images of the three pairs and the result of USO, which has no strength control. Orange frames mark the published defaults of StyleID, StyleGallery, and AD. \method{} keeps the content recognizable at every setting while brushwork and motifs develop step by step. AD and StyleGallery move from the content image to a result with little remaining content within one or two settings, and the effect of their default differs from pair to pair.}
\label{fig:controls:examples}
\end{figure*}

\paragraph{Visual Behavior.}
\Cref{fig:controls:examples} shows three pairs, with one block per method and the same five settings in every row. With \method{}, the sailboat, the owl, and the skyline remain recognizable at all settings while the style develops gradually. Leaf-shaped strokes and then lemons form on the water below the sailboat, stained-glass tracery spreads through the owl's plumage, and the sky and towers break into angular facets. StyleID changes at a similarly even pace over its first three settings, but its last two settings dissolve the content into the texture of the style image. AD and StyleGallery return the content image at their first two settings and lose most of the content at the last two, which leaves the default as the only setting between these extremes. At their strongest settings, the panes of the window replace the owl's face. What the default produces then depends on the image. Both defaults restyle only the hull of the sailboat and the water below it. AD's default then only darkens the owl but has already restyled the skyline heavily, whereas StyleGallery's default keeps the owl close to the input and changes mainly the skyline. Redux changes little up to strength $0.25$. At higher strengths it increasingly replaces the subject with new content drawn from the style image, such as the church window that takes the place of the owl.

\begin{figure*}[t]
\centering
\includegraphics[width=0.97\linewidth]{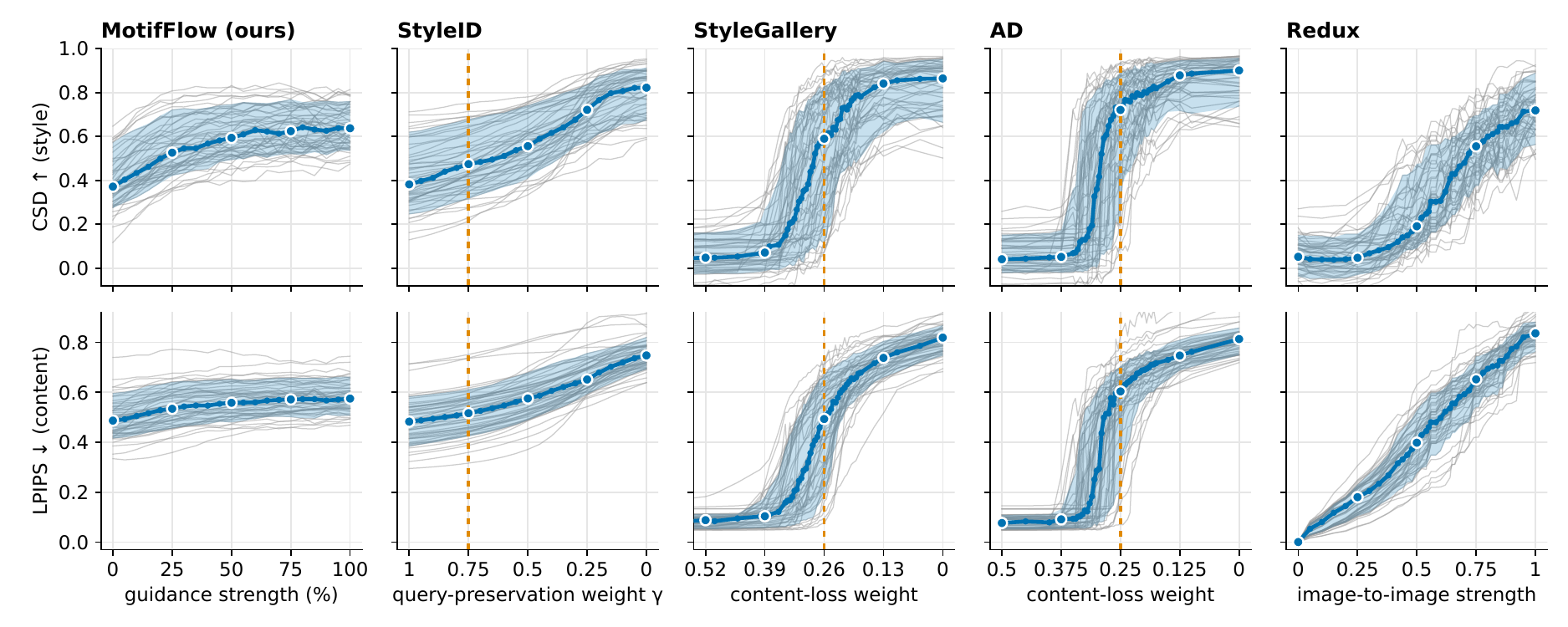}
\caption{\textbf{CSD and LPIPS across each method's strength control for all 50 pairs.} Each gray line follows one content--style pair over a dense sweep of the control. The blue line is the median, with a large marker at each of the five settings and a small marker at every other evaluated value. The band spans the 10th to 90th percentiles, and the dashed vertical line marks the published default of StyleID, StyleGallery, and AD. A narrow band means that a setting changes different pairs by a similar amount. With \method{}, CSD rises from each of the five settings to the next while LPIPS increases only slightly, and its band remains narrow over the whole range. The curves of AD and StyleGallery form a step whose position varies between pairs, so the band covers most of the range throughout the step.}
\label{fig:controls:dist}
\end{figure*}

\begin{table}[t]
\centering\small
\setlength{\tabcolsep}{3.5pt}
\caption{\textbf{Summary of the five-setting comparison.} The CSD and LPIPS columns give the mean over the 50 pairs at the weakest setting, followed by the mean at the strongest setting. The ratio is the increase in LPIPS per unit of CSD gained between these two settings, and the spread is the widest 10th-to-90th-percentile range of LPIPS among the five settings. \method{} gains style at the lowest cost in perceptual change, and its settings act as consistently across pairs as those of StyleID.}
\label{tab:controls}
\begin{tabular}{@{}lccrr@{}}
\toprule
Method & CSD $\uparrow$ & LPIPS $\downarrow$ & Ratio $\downarrow$ & Spread $\downarrow$\\
\midrule
StyleID & .411 $\rightarrow$ .807 & .482 $\rightarrow$ .751 & 0.68 & .20\\
StyleGallery & .063 $\rightarrow$ .829 & .085 $\rightarrow$ .808 & 0.94 & .48\\
AD & .060 $\rightarrow$ .869 & .081 $\rightarrow$ .808 & 0.90 & .38\\
Redux & .060 $\rightarrow$ .724 & .000 $\rightarrow$ .833 & 1.26 & .25\\
\method{} & .386 $\rightarrow$ .643 & .494 $\rightarrow$ .582 & 0.34 & .19\\
\bottomrule
\end{tabular}
\end{table}

\paragraph{Behavior across Pairs.}
\Cref{fig:controls:dist} extends the comparison to all 50 pairs by plotting CSD and LPIPS over the dense sweep of each control with the five settings marked, and \cref{tab:controls} summarizes the five settings. \method{} raises mean CSD from $0.386$ to $0.643$ while mean LPIPS grows only from $0.494$ to $0.582$. At every setting, the 10th and 90th percentiles of LPIPS across pairs differ by at most $0.19$, so a given strength changes different images by a similar amount. StyleID is equally consistent across pairs, but the same gain in style costs it twice the perceptual change. Between its weakest and strongest settings, LPIPS rises by $0.68$ for each unit of CSD gained, compared with $0.34$ for \method{}. AD and StyleGallery behave differently. Their curves stay at the content image for two settings and near the style image for the last two, and the pairs disagree about where the step between the two falls. At AD's default weight, LPIPS ranges from $0.32$ to $0.71$ between the 10th and 90th percentiles, and at StyleGallery's default it ranges from $0.19$ to $0.67$. A single weight therefore leaves some images almost unchanged and removes most of the content from others, and an intermediate result has to be found by tuning the weight for each image. This does not depend on where the five settings are placed. In the dense sweep, each pair moves from the content image to a heavily stylized result within a narrow part of the range, and the position of this transition differs from pair to pair. At every weight whose mean LPIPS falls within the interval covered by \method{}, from $0.49$ to $0.58$, the spread across pairs is at least $0.37$ for AD and $0.27$ for StyleGallery, whereas it is at most $0.19$ at all 21 strengths of \method{}. Among all equally spaced five-setting ranges in the sweep that raise mean CSD by at least $0.20$, the spread in \cref{tab:controls} is never below $0.38$ for AD and $0.42$ for StyleGallery. Redux increases LPIPS almost linearly up to $0.83$, which reflects the replacement of content seen in \cref{fig:controls:examples}. The other methods reach a higher CSD than \method{} at their strongest settings, but they do so at a mean LPIPS above $0.75$, where little of the content image remains. USO offers a single operating point, with mean CSD $0.343$ and LPIPS $0.534$ on these pairs. Its CSD is below that of \method{}'s weakest setting, while its LPIPS lies between those of \method{}'s weakest and strongest settings.

\paragraph{Per-Image Tuning.}
The comparison above applies one set of settings to all images. To also see the baselines at their best, we tune each of them per image. We sweep each baseline's control densely and, for every pair, select the setting with the lowest LPIPS among those whose CSD reaches that of default \method{} within a tolerance of $0.02$. Even when tuned in this way, AD and StyleGallery match rather than surpass the default result of \method{}. They reach its CSD on all 50 pairs at an LPIPS within $0.03$ of that of \method{}, StyleID does so on 49 pairs at an LPIPS within $0.04$, and Redux reaches it on only 40 pairs. These results, however, come from a setting chosen separately for every image. With one weight shared by all images, the results of AD and StyleGallery vary widely from pair to pair, as \cref{fig:controls:dist} and \cref{tab:controls} show. \method{} delivers its result with one strength shared by all images.

\section{Runtime and Backbone Transfer}
\label{supp:additional}

\subsection{Inference Cost}
\label{supp:runtime}

\Cref{tab:runtime} reports median inference times on an H100. Each total covers one content--style pair from start to finish, including the inversion or preprocessing of both images. \method{} processes a pair in $16.5$\,s. Guidance adds $8.3$\,s to injection alone, because each alignment update requires a forward and a backward pass through the transformer. Even with guidance, \method{} is $8.2\times$ faster than \adflux{}, which runs 200 optimization steps on the same backbone. The other methods use the comparison settings of \cref{supp:eval:baselines}.

\begin{table}[t]
\centering\small
\setlength{\tabcolsep}{2.5pt}
\caption{\textbf{Median inference time in seconds for one content--style pair.} Totals include all per-image inversion and preprocessing. Stage times are shown for StyleID and our two configurations.}
\label{tab:runtime}
\begin{tabular}{@{}llrrrr@{}}
\toprule
Method & Backbone & Total & \shortstack{Content\\inversion} & \shortstack{Style\\inversion} & Generation\\
\midrule
StyleID & SD\,1.5 & 6.0 & 2.1 & 2.1 & 1.8\\
StyleGallery & SD\,1.5 & 27.5 & -- & -- & -- \\
AD & SD\,1.5 & 17.8 & -- & -- & -- \\
\midrule
USO & FLUX.1 & 14.1 & -- & -- & -- \\
Redux & FLUX.1 & 5.1 & -- & -- & -- \\
\midrule
\adflux{} & FLUX.1 & 136 & -- & -- & -- \\
Injection only & FLUX.1 & 8.2 & 2.5 & 3.5 & 2.2\\
\method{} & FLUX.1 & 16.5 & 2.5 & 3.5 & 10.5\\
\bottomrule
\end{tabular}
\end{table}

\subsection{FLUX.2 klein}
\label{supp:flux2}

FLUX.2 klein~\cite{bfl2025flux2} natively accepts reference images: their clean latent tokens are appended to the image-token sequence, where the generated image attends to them. On this backbone, we combine attention injection and alignment guidance with this native input, through which we additionally pass the style image. \Cref{tab:flux2} compares this full configuration with the same method without the native style input, injection alone, and injection retaining RoPE. Both guided configurations improve style resemblance and local texture over injection alone. The full method reduces the fraction below CSD $0.30$ from $44.6\%$ to $0.3\%$, showing that the gain includes fewer low-CSD outputs. Retaining RoPE causes the largest perceptual change and greater texture distance, consistent with the need to separate style matching from style-image coordinates.

\begin{table}[t]
  \centering
  \small
  \setlength{\tabcolsep}{3pt}
  \caption{\textbf{Results on FLUX.2 klein.} The last two rows add guidance to injection, and the full method also passes the style image to FLUX.2's native reference-image input. Guidance raises CSD and DINO similarity and lowers texture distance relative to injection alone, while LPIPS increases.}
  \label{tab:flux2}
  \begin{tabular}{@{}l cccc@{}}
    \toprule
    Configuration & LPIPS $\downarrow$ & CSD $\uparrow$ & DINO $\uparrow$ & Texture $\downarrow$ \\
    \midrule
    RoPE kept & 0.800 & 0.567 & 0.578 & 0.619 \\
    Injection only & 0.452 & 0.329 & 0.241 & 0.226 \\
    Without native style input & 0.553 & 0.596 & 0.511 & 0.116 \\
    Full method & 0.579 & 0.631 & 0.554 & 0.094 \\
    \bottomrule
  \end{tabular}
\end{table}

Native style-image conditioning provides an additional source of style information within the same frozen backbone. Compared with the method without this input, it raises style similarity from $0.596$ to $0.631$ and DINO similarity from $0.511$ to $0.554$, while LPIPS increases from $0.553$ to $0.579$. This input therefore strengthens stylization while allowing a larger change to the content.

\section{Limitations and Future Work}
\label{supp:limitations}

\paragraph{Stylization and Content.}
The brushwork and motifs that \method{} transfers reshape the image, so stronger stylization comes with a larger change to the content. Relative to injection alone, guidance raises LPIPS from $0.490$ to $0.560$ in Table~1, and delicate structures such as facial features can be altered. The guidance strength lets a user set this balance, and along this control \method{} gains style at a smaller cost in content than the compared methods do along theirs (\cref{tab:controls}). Strength is, however, the only setting that we vary systematically. Other choices shift the same balance. Restricting injection to the later sampling steps or increasing the weight of the query anchor may preserve more of the content while limiting how far the style can develop. We ablate the query anchor in Sec.~5.5 but do not sweep these settings, and we leave a fuller study of this design space to future work.

\paragraph{Cost.}
Because no module is trained, the style signal is computed during sampling. Each alignment update adds a backward pass through the transformer, so a pair takes $16.5$\,s instead of the $8.2$\,s of injection alone (\cref{tab:runtime}), and both images are inverted beforehand to provide the cached features. \method{} nevertheless remains $8.2\times$ faster than direct optimization on the same backbone, and the style features can be reused across content images. The features could instead be computed during sampling from noised copies of the two images, which would make the cache unnecessary. Obtaining equally stable results without inversion is a direction for future work.

\paragraph{Sparse Styles.}
\method{} transfers the marks that a style image contains, while its content preservation retains the detail of the photograph. Styles defined by what they leave out are therefore harder to reproduce. With pale ink and watercolor, the result can keep too much photographic detail to convey empty space and diffuse washes, and a more faithful rendering of such styles remains a useful direction.

\paragraph{Scope.}
Quantitative evaluation uses $512\times512$ outputs on FLUX.1-dev and FLUX.2 klein, and behavior at higher resolutions and on other backbones requires separate measurement. \method{} is self-contained by design and relies only on the frozen backbone. Combining it with other pretrained modules may open further possibilities, which lie outside the scope of this paper.

\end{document}